\documentclass[letterpaper, 10 pt, conference]{ieeeconf}  

\IEEEoverridecommandlockouts                              

\usepackage{booktabs}
\usepackage{graphicx} 
\usepackage{amsmath}
\usepackage{xcolor}
\usepackage{amssymb}
\usepackage{threeparttable}
\usepackage{multirow}  
\usepackage{siunitx}
\usepackage{float}
\usepackage{dsfont}
\usepackage{subfig}
\usepackage{placeins}
\usepackage{dblfloatfix}
\usepackage{arydshln}
\newcolumntype{Y}{>{\raggedright\arraybackslash}X}

\title{\LARGE \bf
PinPoint3D: Fine-Grained 3D Part Segmentation from a Few Clicks
}

\author{
Bojun Zhang$^{1,2}$, Hangjian Ye$^{1,2}$, Hao Zheng$^{1,2,\dag}$, Jianzheng Huang$^{1}$, Zhengyu Lin$^{1}$, Zhenhong Guo$^{1}$, Feng Zheng$^{1,2,*}$
\thanks{\dag Project Lead.}%
\thanks{$^{1}$All authors are with the Department of Computer Science and Engineering, Southern University of Science and Technology. Emails for the first six authors:
        {\tt\small \{12211615, 12212012, 11610127, 12532586, 12310817, 12312507\}@mail.sustech.edu.cn}}%
\thanks{$^{2}$All authors are with Spatialtemporal AI.}%
\thanks{*Corresponding author: {\tt\small zfeng02@gmail.com}}
}

\begin{document}

\maketitle
\thispagestyle{empty}
\pagestyle{empty}

\begin{abstract}

Fine-grained 3D part segmentation is crucial for enabling embodied AI systems to perform complex manipulation tasks, such as interacting with specific functional components of an object. However, existing interactive segmentation methods are largely confined to coarse, instance-level targets, while non-interactive approaches struggle with sparse, real-world scans and suffer from a severe lack of annotated data. To address these limitations, we introduce PinPoint3D, a novel interactive framework for fine-grained, multi-granularity 3D segmentation, capable of generating precise part-level masks from only a few user point clicks. A key component of our work is a new 3D data synthesis pipeline that we developed to create a large-scale, scene-level dataset with dense part annotations, overcoming a critical bottleneck that has hindered progress in this field. Through comprehensive experiments and user studies, we demonstrate that our method significantly outperforms existing approaches, achieving an average IoU of around 55.8\% on each object part under first-click settings and surpassing 71.3\% IoU with only a few additional clicks. Compared to current state-of-the-art baselines, PinPoint3D yields up to a 16\% improvement in IoU and precision, highlighting its effectiveness on challenging, sparse point clouds with high efficiency. Our work represents a significant step towards more nuanced and precise machine perception and interaction in complex 3D environments.

\end{abstract}

\section{INTRODUCTION}

The advancement of embodied AI, from household assistants to industrial robots, is critically dependent on the ability to perceive and interact with the world at a human-like level of detail. Consider a service robot tasked with retrieving an item from a specific drawer in a cabinet. An instance-level understanding \textit{``recognizing the cabinet as a whole''} is insufficient. The robot must identify and act upon a particular functional \textit{part} (the drawer handle) among many. This requires a fine-grained, hierarchical understanding of the scene, progressing from the general environment to specific action-relevant components (i.e., scene $\rightarrow$ area $\rightarrow$ instance $\rightarrow$ part). While the need is clear, acquiring dense, part-level 3D annotations required to train such systems remains a formidable bottleneck due to prohibitive labor and scaling costs.

\begin{figure}[t!]
    \centering
    \includegraphics[width=0.9\columnwidth]{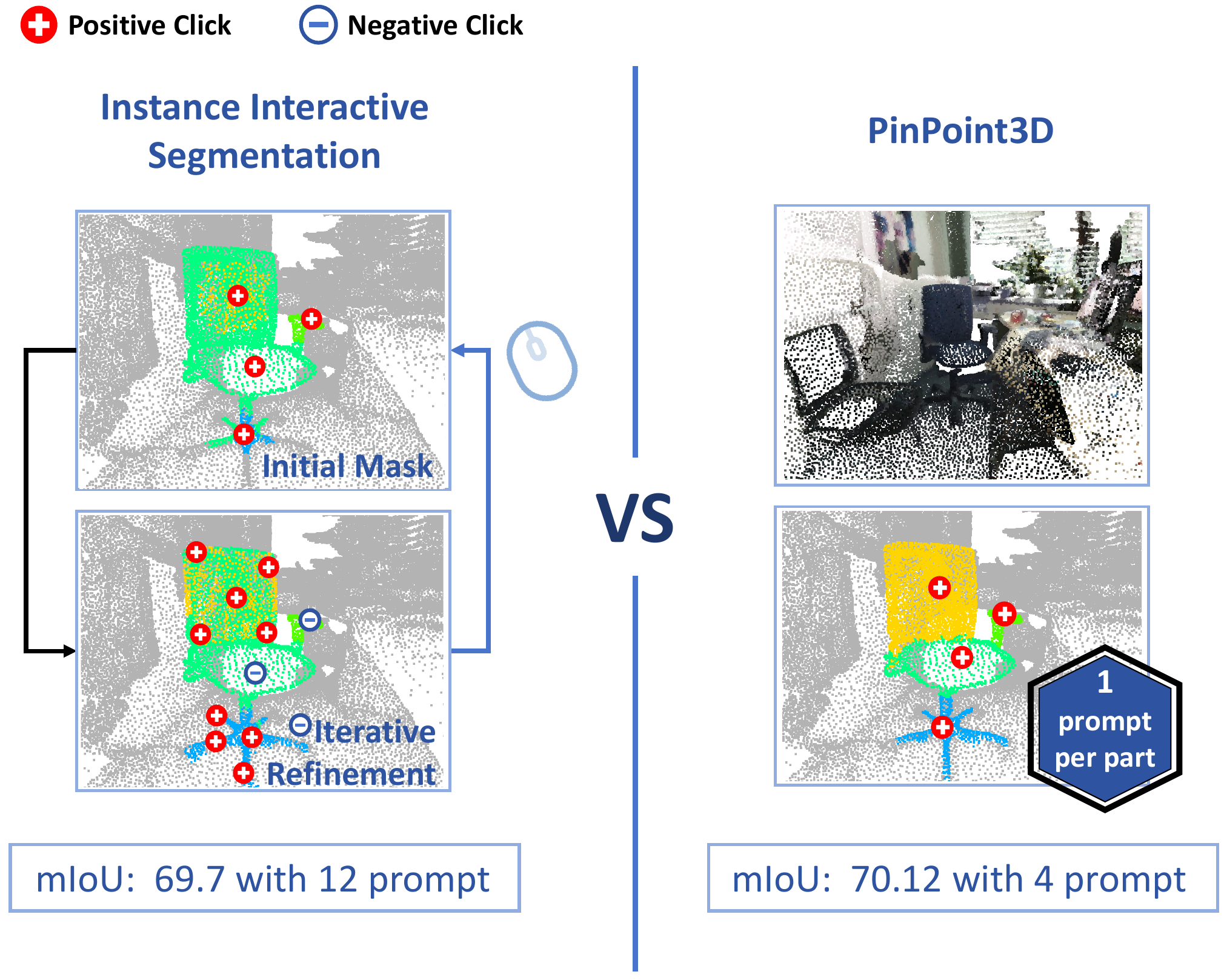}
    \caption{
        \textbf{Purpose-Built Interactive Part Segmentation.} Instance-level models repurposed for part segmentation (\textit{left}) require extensive user interaction (e.g., 12 clicks). In contrast, our purpose-built framework (\textit{right}) delivers superior accuracy with minimal effort, requiring only a single click per part.
    }
    \label{fig:intro_comparison}
\end{figure}

Interactive 3D segmentation has emerged as a promising solution to reduce annotation burden~\cite{Kontogianni2023, Yue2023, Sun2023, Zhang2024, zhou2025pointsam}. However, existing methods primarily target coarse, instance-level segmentation. As illustrated in Figure~\ref{fig:intro_comparison}, applying these models to part-level tasks is inefficient and often yields inaccurate results. Current approaches that attempt interactive part segmentation face other significant limitations: they either process single objects in isolation~\cite{lang2024iseg} or depend on 2D foundation models that create information bottlenecks, failing to capture essential 3D geometric details~\cite{zhou2025pointsam}. Moreover, non-interactive part segmentation and hierarchical methods, while effective on clean CAD models, suffer substantial performance degradation when applied to sparse, noisy point clouds from real-world scans~\cite{Hsu2022, Fan2022, Li2022}. These limitations underscore the critical need for an interactive framework capable of robust part segmentation in challenging scene-level data.

We present PinPoint3D, a novel interactive framework designed for fine-grained, multi-granularity 3D segmentation. Our approach excels at processing sparse, scene-level point clouds, generating highly accurate part masks with minimal user input. The framework operates through several coordinated stages. First, we employ a sparse convolutional network to extract point features, utilizing a lightweight adapter module to align feature dimensions with the downstream decoder. User-provided point prompts are encoded as learnable query features, enhanced with spatial-temporal embeddings that preserve both the relative positions and sequence of user interactions. These encoded queries guide a transformer decoder to generate object-level masks within the scene.

Simultaneously, a dedicated part-level branch focuses on individual objects, employing a specialized decoder to segment fine-grained parts. We introduce an attention mechanism that facilitates information exchange between part components and enables communication between parts and their parent instances. This hierarchical decoding strategy is supported by an iterative refinement loop with dynamic attention masking, where the predicted masks from each decoding step inform subsequent attention focus areas. Within this framework, objects are treated as miniature scenes where part components are extracted, mirroring how instance segmentation extracts objects from larger scenes. The final mask prediction module generates segmentation results for both instance and part queries. Throughout the iterative process, the background query excludes points claimed by parts, while part queries progressively delineate distinct object segments.

Our experimental evaluation across three diverse 3D point cloud datasets demonstrates that PinPoint3D achieves 55.8\% average IoU with just a single positive click per target part, exceeding 71.3\% IoU with minimal additional user input. Compared to existing interactive segmentation methods, our framework substantially reduces required user effort while improving segmentation accuracy.

Our primary contributions include:

\begin{itemize}
    \item A novel interactive segmentation framework that efficiently generates multi-granularity masks from sparse point-based user inputs, bridging the gap between instance and part-level understanding.
    
    \item A comprehensive 3D data synthesis pipeline, along with a large-scale, scene-level dataset with dense part annotations constructed by this pipeline, addressing the critical data scarcity in this domain.
    
    \item An intuitive interactive mechanism and user interface, validated through extensive user studies, demonstrating superior usability and performance compared to existing fine-grained segmentation approaches.
    
    \item Robust experimental validation across multiple datasets and unseen scenes, confirming our method's generalization capabilities and practical applicability.
\end{itemize}

\section{RELATED WORKS}

\subsection{3D Interactive Segmentation} 
There are only a few methods that support interactive 3D segmentation with explicit user input. Valentin et al. \cite{valentin2015semanticpaint} and Zhi et al. \cite{zhi2022ilabel} introduced early systems for online 3D scene labeling, focusing on semantic annotation rather than instance segmentation. Shen et al. \cite{shen2020scribbles} projected the interaction onto the 2D domain by allowing users to annotate multi-view images with known camera poses through scribbles, but providing feedback from multiple viewpoints is cumbersome. More recent approaches operate directly on 3D point clouds: Kontogianni et al. \cite{Kontogianni2023} proposed InterObject3D, which lets users click on a point cloud to segment objects but can only handle a single object each time. Yue et al. \cite{yue2024AGILE3D} developed an attention-guided model that segments multiple objects simultaneously by encoding user clicks as spatial queries, achieving higher accuracy with fewer clicks. Zhou et al. \cite{zhou2025pointsam} adapts the Segment-Anything concept to 3D point clouds for promptable segmentation and demonstrates strong generalization across domains. For fine-grained part-level segmentation on individual 3D shapes, Lang et al. \cite{lang2024iseg} present an interactive mesh segmentation technique that processes positive and negative clicks on a shape's surface to produce a binary part mask.

\subsection{Multi-granularity segmentation}
Current research on 3D segmentation has largely progressed on scene-level representations that enable multi-scale structure driven by language or interaction. Distillation approaches such as OpenScene \cite{opnscene} and ConceptFusion\cite{conceptfusion} yield sufficiently fine features but lack explicit hierarchical structure and controllable granularity. Relatedly, implicit scene representations, including LERF\cite{lerf} and Open-NeRF\cite{opennerf}, offer queryable language embeddings and decomposition capabilities, yet operate mainly through \emph{implicit} field correlations and still do not provide explicit hierarchy/granularity control.

Recent work injects \emph{controllable granularity} or \emph{hierarchical} information at the representation level. N2F2\cite{n2f2} encodes nested feature dimensions to support querying the same scene at multiple granularities within a single field; GARField\cite{garfield} represents scene elements using an affinity field and directly adjusts grouping granularity via an extra control parameter; SAMPart3D\cite{sampart3d} introduces \emph{scale-conditioned} part-aware features to achieve multi-granularity part segmentation without a predefined part vocabulary. A series of works lifts 2D SAM~\cite{kirillov2023segment} 's multi-granularity masks into 3D: SA3D and SAI3D rely on multi-view consistency and region growing to obtain instance/part masks, while SAGA employs \emph{scale-gated affinity} features for promptable and granularity-adjustable 3D segmentation\cite{sam3d,sai3d,saga3d,huang2024segment3d}.

\subsection{3D Part Segmentation}

3D part segmentation involves two main tasks: semantic segmentation and instance segmentation. Many 3D networks~\cite{qi2017pointnetpp, qian2022pointnext, thomas2019kpconv, wang2019dgcnn} predict a semantic label for each point or voxel in a 3D shape. Existing learning-based approaches tackle instance segmentation by incorporating various point grouping and clustering strategies~\cite{chu2021icm, he2020prototypes, jiang2020pointgroup, liu2020selfseg, vu2022softgroup, wang2018sgpn, wang2019asis, zhang2021probabilistic} or by generating part proposals with 3D bounding boxes and region proposal networks~\cite{hou2019sis, yang2019boenet, yi2019gspn} in the pipeline. Some methods leverage weak supervision for 3D part labeling, using coarse annotations such as bounding boxes~\cite{chibane2022box2mask, liu2022box2seg}, language reference games~\cite{koo2022partglot}, or assembly manuals~\cite{wang2022ikeamanual} instead of dense labels. 
Instead of focusing only on individual objects, a series of methods perform part segmentation on 3D scenes~\cite{bokhovkin2021rgbd, notchenko2022scan2part}, identifying object parts within large RGB-D scans. Another line of work~\cite{luo2020partdiscover, wang2021nopartlabels, yu2019PartNet} decomposes 3D shapes into a set of fine-grained (often hierarchical) parts without using semantic labels. 

Recently, the emergence of vision-language and other foundation models has enabled 3D part segmentation with minimal manual labels. PartSLIP~\cite{liu2023partslip} transfers knowledge from a pretrained 2D image-language model to detect part regions in multi-view renderings of a 3D object, achieving open-vocabulary zero-shot part segmentation via 2D-to-3D label lifting. PartSTAD~\cite{kim2024partstad} adapts 2D SAM~\cite{kirillov2023segment} to 3D point clouds via few-shot training, improving part segmentation by leveraging 2D pretrained priors. SAMPart3D~\cite{yang2024sampart3d} distills a vision foundation model into a 3D feature backbone, enabling zero-shot part segmentation without predefined part prompts.

\section{METHOD}

\begin{figure}[t!]
    \centering
    \includegraphics[width=0.7\columnwidth]{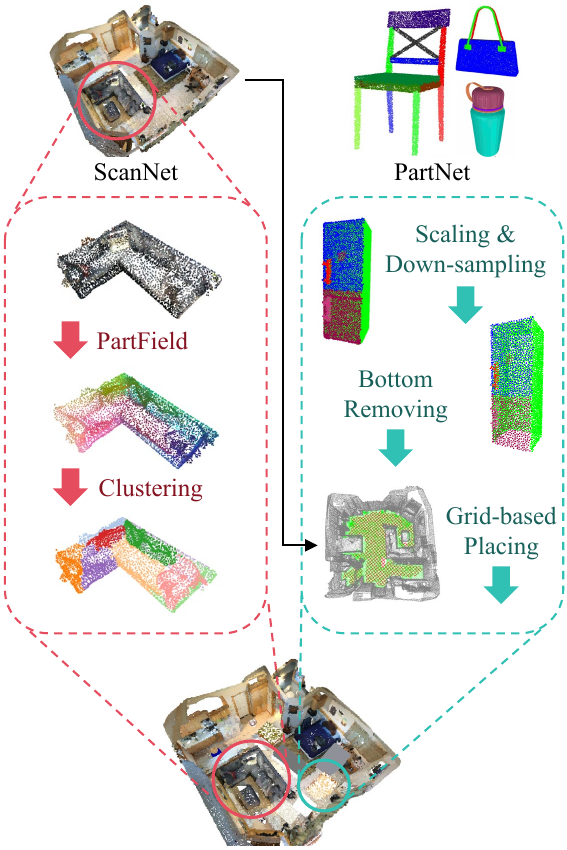}
    \caption{
        \textbf{The process of dataset construction} \textit{\textbf{left}}:generate Pseudo Labels.Using PartField model to extract feature of ScannNet and clustering to get part annotations \textit{\textbf{right}}:Synthetic Data Generation. Adapting PartNet to ScanNet
    }
    \label{fig:data}
\end{figure}

\subsection{Overall Architecture}
We propose a novel \textit{TWO-Level Forward Mask Prediction Architecture} that simultaneously produces object-level and part-level segmentation masks within a unified decoder pipeline, as illustrated in Fig~\ref{fig:archi}. Compared to existing interactive 3D segmentation methods, our framework is specifically designed to \textbf{(1) frozen backbone with lightweight adapter}, which preserves stable object-level semantics while enabling fine-grained adaptation to part-level segmentation, and \textbf{(2) dedicated part-level decoder}, which builds upon object-level predictions and refinement into a fine-grained part mask. 

In order to enhance \textbf{hierarchical consistency}, we propose the idea of a Targeted Attention Mask(TAM) that constrains part queries to operate within the target object. In practice, this mechanism can be applied to prevent cross-object interference, while its principle is implicitly embedded in the training by simulating part clicks into single-object regions.


\subsection{Data Generation}
\label{sec:data-gen}
Our approach requires a large scale of point cloud with part annotations for training. Due to the scarcity of such annotated datasets, we construct a novel training set by integrating data from ScanNet and PartNet. Figure~\ref{fig:data} illustrates the process of dataset construction. The first step is to generate part-level pseudo-annotations for objects in the ScanNet scenes. Then we build our synthesis pipeline that embeds fine-grained object meshes from PartNet into ScanNet scenes.

\subsubsection{Generate Pseudo Labels} 
ScanNet is an indoor scene dataset offering 3D point clouds with instance-level annotations~\cite{dai2017ScanNet}. For each object instance that can be decomposed into semantic parts, we augment it by generating part-level pseudo-labels using PartField~\cite{liu2025partfield}. PartField captures the general concept of parts and their hierarchical structure, generating a continuous feature field for a given object. These learned point-wise representations are then clustered to produce part segmentations. This procedure yields a part segmentation that is both compact and well-separated, aligning with the object's intrinsic structure.

\begin{figure*}[t]
    \centering
    \includegraphics[width=\textwidth]{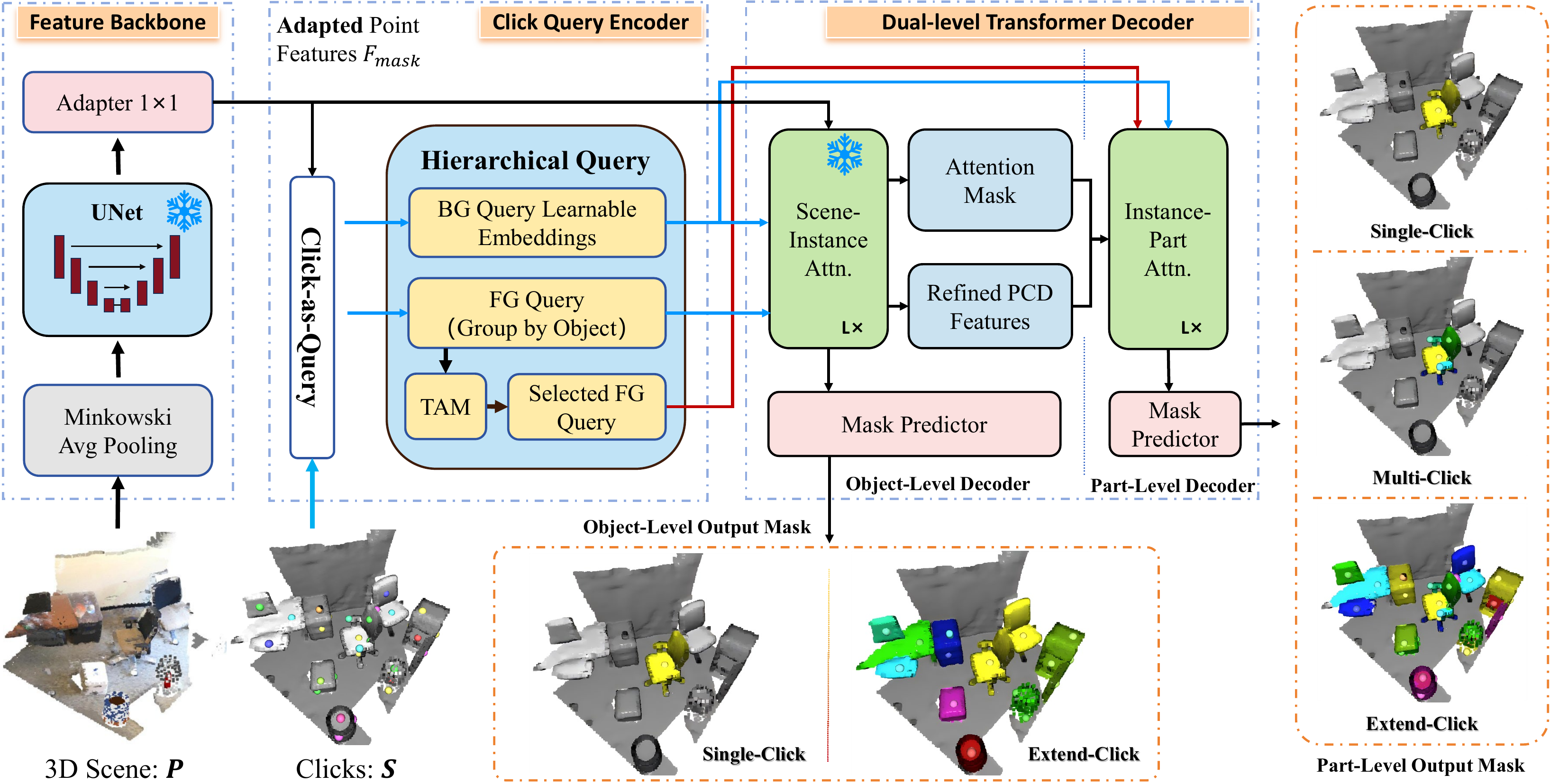} 
    \caption{Hierarchical interactive segmentation pipeline. Given a 3D scene $P$ and user clicks $S$, the \textbf{Feature Backbone} (Minkowski U-Net with a $1{\times}1$ Adapter) extracts per-point features. The \textbf{Click Query Encoder} forms hierarchical queries (learnable background embeddings; object-grouped foreground queries) and TAM selects active queries. The \textbf{Dual-level Transformer Decoder} refines features via Scene--Instance and Instance--Part attention to predict masks. Two heads are provided: an \textbf{optional Object-Level Decoder} for holistic masks, and a \textbf{Part-Level Decoder} that is trained for fine-grained part segmentation while remaining object-consistent in low-click regimes, thereby yielding object-level masks when required—even without invoking the optional object head.}
    \label{fig:archi}
\end{figure*}

\subsubsection{Synthetic Data Generation}
To overcome the potential imprecision of pseudo-labels derived from ScanNet, we augment our training data with high-quality assets from the PartNet dataset~\cite{mo2019partnet}. PartNet provides fine-grained, hierarchically consistent part annotations for a diverse range of 3D objects. This hybrid data strategy allows our model to learn from both the high-fidelity part supervision of PartNet and the complex spatial arrangements of real-world indoor scenes from ScanNet, which is crucial for enhancing its generalization capabilities.

\begin{table}[htbp]
\centering
\small
\caption{Per-category IoU at different thresholds}
\label{tab:per-category-iou}
\renewcommand{\arraystretch}{1.2} %
\setlength{\tabcolsep}{7pt} %
\begin{tabular}{lcccc}
\toprule
Category & $\text{IoU}_{1}$ & $\text{IoU}_{3}$ & $\text{IoU}_{5}$ & mIoU \\
\midrule
Bag              & 53.8 & 63.4 & 68.1 & 61.8 \\
Bottle           & 62.0 & 67.0 & 68.3 & 65.8 \\
Chair            & 62.4 & 76.7 & 84.2 & 74.4 \\
Dishwasher       & 47.2 & 62.9 & 67.1 & 59.1 \\
Faucet           & 55.7 & 69.6 & 70.8 & 65.4 \\
Lamp             & 52.3 & 66.7 & 70.3 & 63.1 \\
Microwave        & 34.7 & 54.2 & 59.8 & 49.6 \\
Mug              & 66.5 & 74.0 & 75.2 & 71.9 \\
Refrigerator     & 51.2 & 65.1 & 70.5 & 62.3 \\
StorageFurniture & 60.8 & 76.0 & 79.7 & 72.2 \\
Table            & 65.2 & 76.8 & 80.4 & 74.1 \\
Vase             & 58.1 & 69.1 & 72.2 & 66.5 \\
\midrule
\bottomrule
\end{tabular}
\end{table}

Our data integration process involves carefully selecting and placing PartNet objects into ScanNet scenes. To align the selected PartNet objects with the scale and density of ScanNet scenes, we first analyzed key statistical properties, including size, volume, and point density across both datasets. 

Based on these statistics, we determined an appropriate scaling factor for each PartNet category to match the geometric properties of ScanNet counterparts. We then scaled and downsampled each PartNet object point cloud accordingly. 
Each object was scaled and downsampled using farthest point sampling, applied independently to each annotated part. This preserves local structure and part-level detail.

The processed PartNet objects were then inserted into ScanNet scenes to create synthetic indoor environments enriched with accurate part annotations. Specifically, for a given ScanNet scene, we first estimate the floor plane and partition it into a 2D grid. Objects are randomly placed into unoccupied grid cells, one at a time, until either the available space is exhausted or a predefined object limit is reached. This procedure yields augmented 3D scenes that contain both the original ScanNet geometry with pseudo-labeled parts and PartNet objects with accurate part annotations.
\subsection{Model Architecture}

Our framework is an attention-based two-stage 3D segmentation model that takes a sparse point cloud and a few user-provided point prompts as input, and produces fine-grained part segmentation masks. 
It consists of (a) a 3D sparse convolutional \textbf{backbone} for feature extraction, (b) a \textbf{click query encoder} that generates learnable query embeddings from user clicks, and (c) a dual-level transformer \textbf{decoder} with Targeted Attention Masking (TAM) for instance and part decoding.

\subsubsection{Feature Backbone}

We adopt a 3D sparse convolutional backbone~\cite{minkowski-engine}  for point cloud feature extraction. Operating on a sparse voxel grid, it efficiently encodes the $N$ input points into a feature tensor $\mathbf{F}_{\text{pcd}} \in \mathbb{R}^{N \times C_{\text{bb}}}$ with rich geometric context. To adapt the feature backbone for fine-grained 3D segmentation tasks, we insert a lightweight \emph{residual adapter} that specializes the object-level representations for \emph{part-level} semantics. Inspired by He et al.~\cite{he-etal-2021-effectiveness}, the adapter applies a bottleneck-style update $\Delta(\cdot)$ and a controlled residual scaling.
Formally, given the backbone features $\mathbf{F}_{\text{pcd}}$, the adapter produces 
\begin{equation}
\small
\mathbf{F}_{\text{mask}} \;=\; \mathbf{F}_{\text{pcd}} \;+\; \alpha \cdot \text{Conv}_{1\times1}^{2}\Big(\text{ReLU}\big(\text{Conv}_{1\times1}^{1}(\mathbf{F}_{\text{pcd}})\big)\Big)\,,
\end{equation}
where $\mathrm{Conv}_{1\times1}^{1}$ reduces channels and $\mathrm{Conv}_{1\times1}^{2}$ expands them; $\alpha\!\in\!(0,1]$ stabilizes residual updates. This design keeps object-level geometry while endowing $\mathbf{F}\!\in\!\mathbb{R}^{N\times D}$ with part-sensitive cues for decoding.

\subsubsection{Click Query Encoder}

We encode the user's point prompts (clicks) as a set of learnable query embeddings. We encode each click using Fourier positional encoding~\cite{tancik2020fourier}. Furthermore, for interactive settings where clicks are provided sequentially, we add a 1-dimensional temporal encoding to each query, indicating the click order. After the encoding steps, queries are sent into a dual-level transformer decoder module for further refinement. 

\subsubsection{Dual-level Transformer Decoder}

We design a dual-level decoder that contains a Scene-Instance decoder, a Targeted Attention Masking(TAM) module, and an Instance-Part decoder. This transformer decoder takes the scene features and user query embeddings as input, producing refined representations for downstream mask prediction at both instance and part levels.

\paragraph{Scene-Instance decoder}
The Scene-Instance decoder processes the foreground and background queries $C_f$ and $C_b$, together with the input scene feature $F_{pcd}$, to obtain updated query embeddings and updated point cloud features. A Scene-Instance Attention Module allows \emph{bidirectional} inter between point prompt queries and point cloud features, as in AGILE3D~\cite{Yue2023}. Attention flows from instance queries into scene features to gather contextual information, then conversely flows from scene features back to instance queries to update the scene representation. Finally, attention is applied among instance queries for direct information exchange.

\paragraph{Mask Prediction and Targeted Attention Masking(TAM)} 
The refined queries generated by the Scene-Instance decoder are fed into a mask prediction module to produce binary masks for each object query. This module uses a learned MLP to map the query embedding vectors to scalar mask logits for each point. Specifically, let $\{\mathbf{w}_i\}$ denote the learned mask embeddings (one for each query). The logit for query $i$ at point $p$ is computed as the dot product between the point's feature $\mathbf{F}_{\text{mask}}(p)$ and the query's embedding $\mathbf{w}_i$: \[z_{i,p} = \mathbf{F}_{\text{mask}}(p)^\top \mathbf{w}_i.\]These logits are then converted to per-point class assignments via a $\max$ operation.  

Based on the initial object mask predictions, we introduce a \textbf{Targeted Attention Masking (TAM)} module to refine the attention maps by focusing on each instance's interior. TAM constructs attention masks from the object predictions and restricts subsequent self-/cross-attention to instance interiors, explicitly encoding the hierarchy between objects and their parts and guiding the downstream Instance-Part decoder.

Given object-level predictions $\hat{y}_n \in \{0,\dots,M\}$, we derive a binary mask 
$\mathbf{A}^{(t)} \in \{0,1\}^{Q\times N}$ for target instance $t$, 
where $\mathbf{A}^{(t)}_{q,n}=0$ if query $q$ is allowed to attend to point $n$, and $1$ otherwise. 
During part decoding, query–point attention is restricted as
\begin{equation}
\alpha_{q,n} \;=\; 
\frac{\exp(s_{q,n}) \cdot \mathds{1}[\mathbf{A}^{(t)}_{q,n}=0]}
{\sum_{n': \mathbf{A}^{(t)}_{q,n'}=0} \exp(s_{q,n'})},
\end{equation} 

This module encodes the object$\rightarrow$part hierarchy: foreground queries for object $t$ attend only to its interior, while background queries attend only to background regions. In practice, $\mathbf{A}^{t}$ is passed as the transformer \textit{memory mask} between the object-level decoder and the part-level decoder (Fig.~\ref{fig:archi}).

\paragraph{Instance-Part Decoder}
For the selected object $i$, a variable number of part queries $\mathbf{P}^{i}\!\in\!\mathbb{R}^{N_p^{i}\times d}$ interact with the object-restricted point set $\mathbf{O}^{i}$ (points gated by TAM). Each layer applies masked cross-attention and self-attention, followed by intra-part self-attention and FFNs with residual connections and normalization. The decoder supports iterative refinement: users can add clicks to update $\mathbf{P}^{i}$ and re-run masked attention, yielding refined part hypotheses $\mathbf{P}^{i}_{\text{final}}$ and precise part masks.

\subsection{Training and Evaluation Protocol}
\subsubsection{Iterative Multi-Part Training Protocol}
We follow the interactive training paradigm of AGILE3D~\cite{Yue2023}, where users iteratively provide simulated clicks. At each refinement step, positive clicks are sampled from parts of the selected objects, while negative clicks are drawn from background regions. Unlike protocols restricted to a single object, we simulate a more realistic scene-level interaction by randomly selecting parts across all objects present in the scene. The number of active parts per interaction is capped at 10, ensuring that each forward pass maintains a tractable scope while still covering diverse object-part configurations. This random yet bounded sampling strategy promotes balanced supervision across scenes of varying complexity, while preserving the realism of simulated annotation. As in AGILE3D, gradients are only backpropagated at the final refinement step to reduce computational overhead.

\subsubsection{Click Simulation for Scene-Level Parts}
In contrast to object-centric settings, our framework simulates part clicks at the scene level. 
For evaluation, we construct two validation protocols for each object: 
(i) \textit{multi-part}, where a random number of parts between 1 and the maximum available parts are annotated; and 
(ii) \textit{full-part}, where all parts of the object are annotated. 
This design allows us to assess model performance under different annotation budgets, ranging from minimal user input to complete coverage.
\subsubsection{Optimization Objective}
Both object- and part-level predictions are supervised by a combination of cross-entropy and Dice loss, encouraging precise boundary alignment in cases of class imbalance. During training, the backbone network remains frozen while the adapter and two decoders are being updated. 
\begin{equation}
\mathcal{L}_{\text{part}}
= \frac{1}{N_{\text{part}}} \sum_{p \in P_{\text{part}}} w_p \,
   \big(\lambda_{\text{CE}} \, \mathcal{L}_{\text{CE}}(p)
   + \lambda_{\text{Dice}} \, \mathcal{L}_{\text{Dice}}(p)\big),
\end{equation}
This stabilizes object-level semantics while enabling fine-grained adaptation to part-level segmentation.

\begin{table*}[t]
\vspace*{5pt}
\centering
\small
\caption{Comparison of Methods on Scene Part Segmentation Ability.}
\label{tab:eval-scene-part-seg}
\renewcommand{\arraystretch}{1.1} %
\setlength{\tabcolsep}{4.5pt} %
\begin{tabular}{l c c c c c c c c c}
\toprule
\textbf{Method} & \textbf{Eval} & $\text{IoU}_{1}\uparrow$ & $\text{IoU}_{3}\uparrow$ & $\text{IoU}_{5}\uparrow$ & $\text{NoC}_{50}\downarrow$ & $\text{NoC}_{65}\downarrow$ & $\text{NoC}_{80}\downarrow$ & $\text{AP}_{25\%}\uparrow$ & $\text{AP}_{50\%}\uparrow$ \\

\midrule
PointSAM & \multirow{3}{*}{SyntheticData(random-part)} & 46.2 & 50.1 & 51.4 & - & - & - & 72.8 & 49.9 \\
AGile3D & & 39.8 & 58.4 & 64.9 & 3.07 & 5.38 & 7.89 & 94.4 & 73.5\\
\textbf{PinPoint3D(Ours)} & & \textbf{50.0} & \textbf{65.9} & \textbf{69.7} & \textbf{2.12} & \textbf{3.92} & \textbf{6.92} &\textbf{95.1} & \textbf{81.5}\\
\midrule
PointSAM &  \multirow{3}{*}{SyntheticData(all-part)} & 48.4 & 52.6 & 52.7 & - & - & - & 74.1 & 51.0 \\
AGile3D & & 39.1 & 61.1 & 66.7 & 2.67 & 5.18 & 8.12 &96.7 &78.2 \\
\textbf{PinPoint3D(Ours)} & & \textbf{55.8} & \textbf{68.4} & \textbf{71.3} & \textbf{1.68} & \textbf{3.46} & \textbf{6.43} & \textbf{96.9} & \textbf{85.7}\\


\midrule
PointSAM &  \multirow{2}{*}{MultiScan(random-part)} & \textbf{44.4} & 54.9 & 58.1 & - & - & - & 81.7 & 57.6\\

AGile3D & & 40.8 & 59.3 & 66.5 & 2.88 & 5.24 & 7.75 & 93.4 & 74.9\\
\textbf{PinPoint3D(Ours)} & & 44.0 & \textbf{60.8} & \textbf{66.8} & \textbf{2.71} & \textbf{4.93} & \textbf{7.74} & \textbf{94.0} & \textbf{77.3} \\

\midrule

PointSAM & \multirow{2}{*}{MultiScan(all-part)} & \textbf{44.9} & 54.0 & 56.1 & - & - & - & 80.8 & 58.3\\

AGile3D & & 42.1 & 61.2 & 67.5 & 2.62 & 4.90 & 7.85 & 93.7 & 76.5 \\
\textbf{PinPoint3D(Ours)} & & 44.4 & \textbf{62.7} & \textbf{68.1} & \textbf{2.28} & \textbf{4.53} & \textbf{7.66} & \textbf{93.9} & \textbf{78.9}\\

\bottomrule
\end{tabular}
\label{tab:cross_dataset_full}
\end{table*}

\begin{figure*}[t]
\centering
\small
\setlength{\tabcolsep}{1pt} %
\renewcommand{\arraystretch}{0.7} %

\makebox[\textwidth][c]{%
\begin{tabular}{cc|ccc|ccc|c}

    \multirow{3}{*}{\rotatebox{90}{PartScan}} &
    \includegraphics[width=0.11\textwidth]{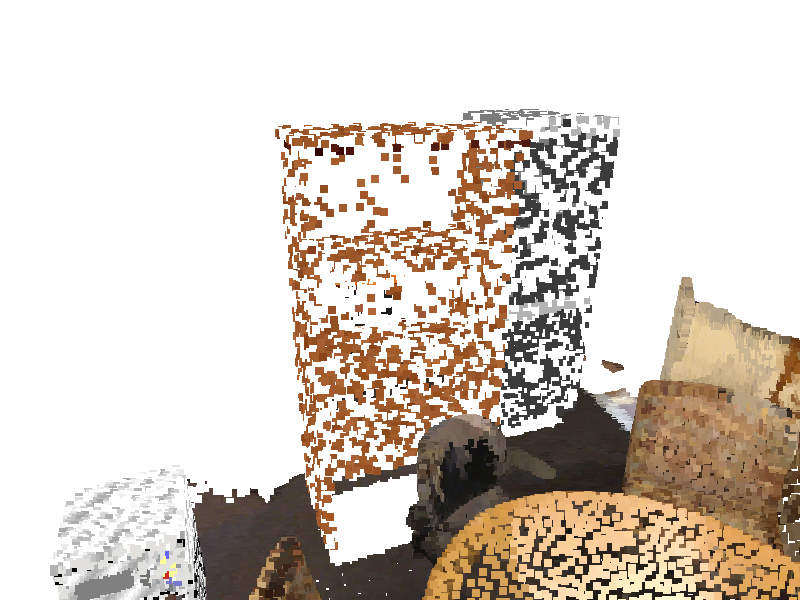} &
    \includegraphics[width=0.11\textwidth]{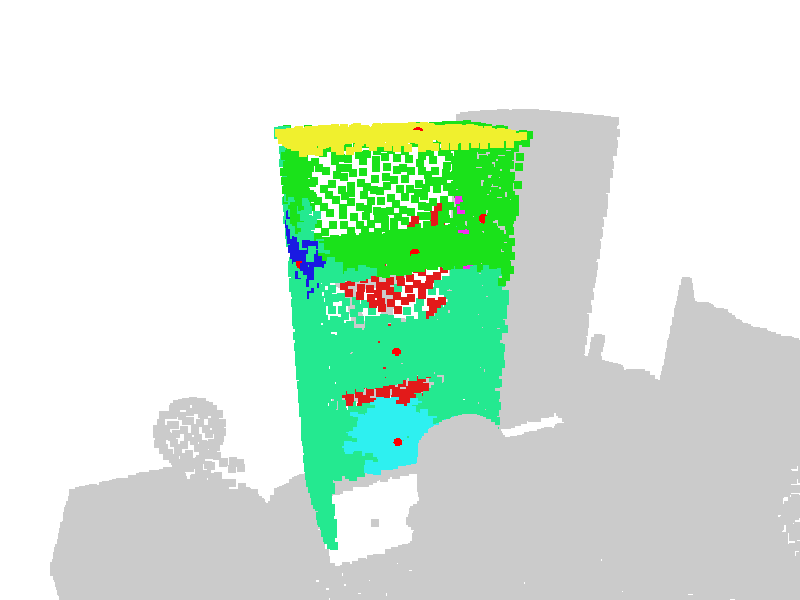} &
    \includegraphics[width=0.11\textwidth]{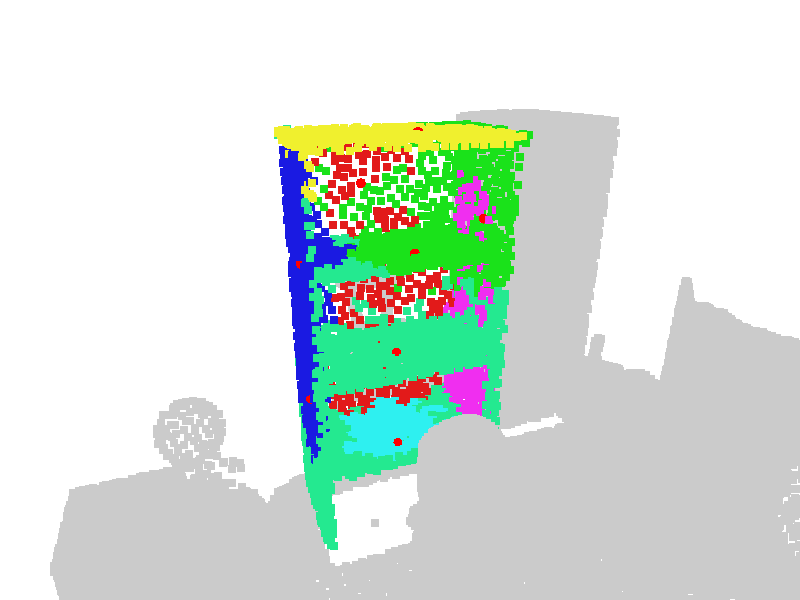} &
    \includegraphics[width=0.11\textwidth]{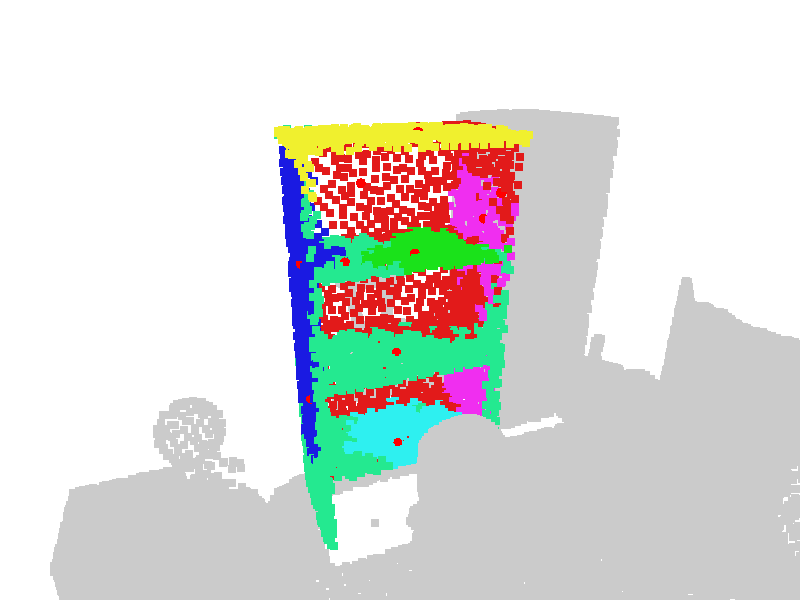} &
    \includegraphics[width=0.11\textwidth]{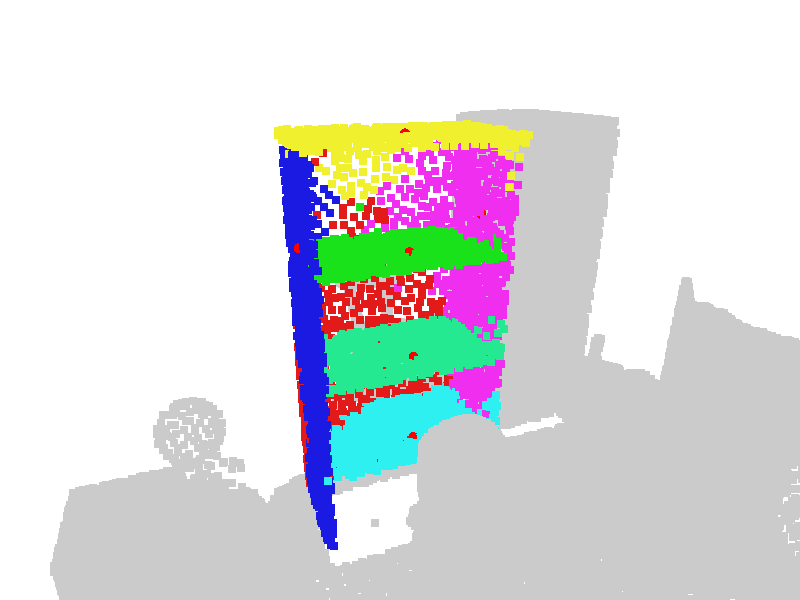} &
    \includegraphics[width=0.11\textwidth]{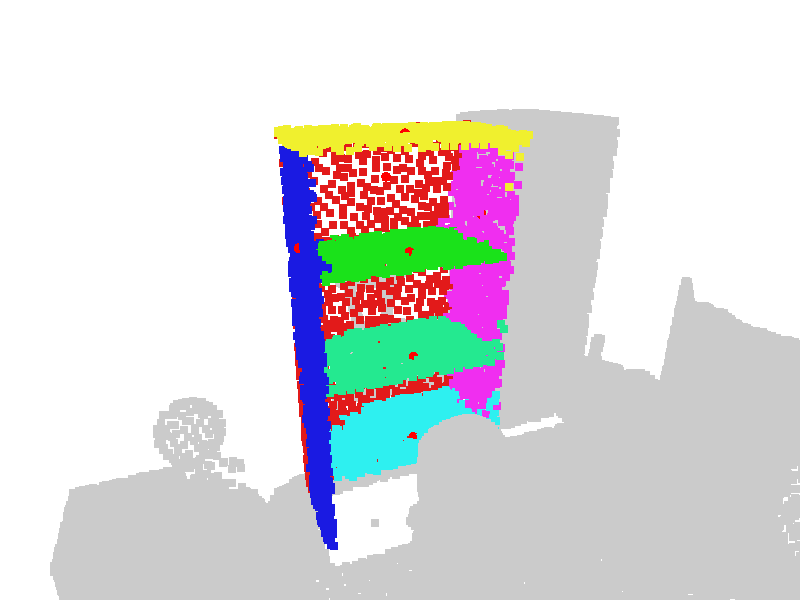} &
    \includegraphics[width=0.11\textwidth]{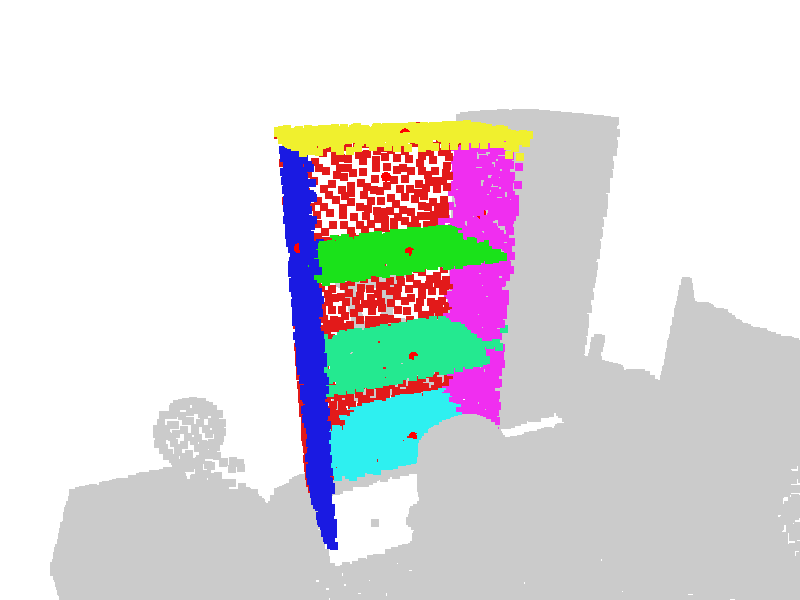} &
    \includegraphics[width=0.11\textwidth]{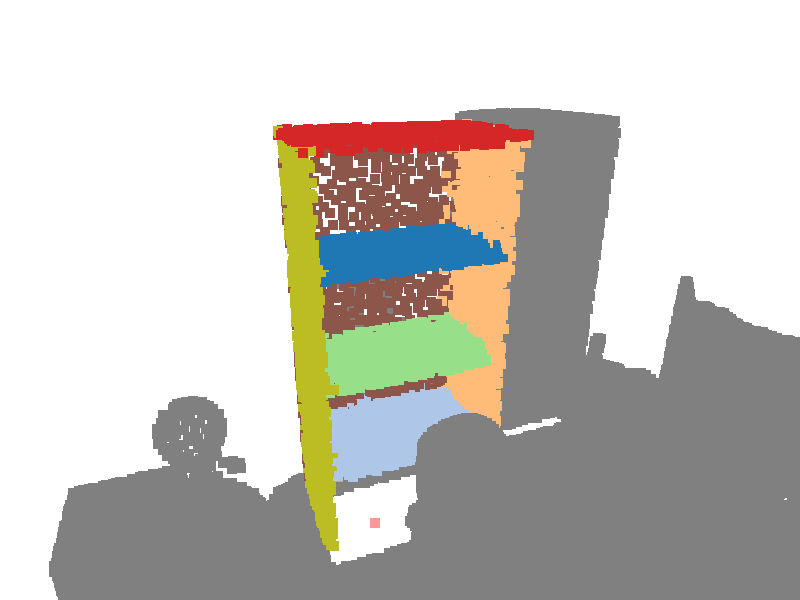} \\
    & & {\scriptsize $\overline{\text{IoU}}@7=32.3$} & {\scriptsize $\overline{\text{IoU}}@10=46.7$} & {\scriptsize $\overline{\text{IoU}}@14=53.2$} & {\scriptsize $\overline{\text{IoU}}@7=75.3$} & {\scriptsize $\overline{\text{IoU}}@8=83.6$} & {\scriptsize $\overline{\text{IoU}}@9=86.1$} & \\

    & \includegraphics[width=0.11\textwidth]{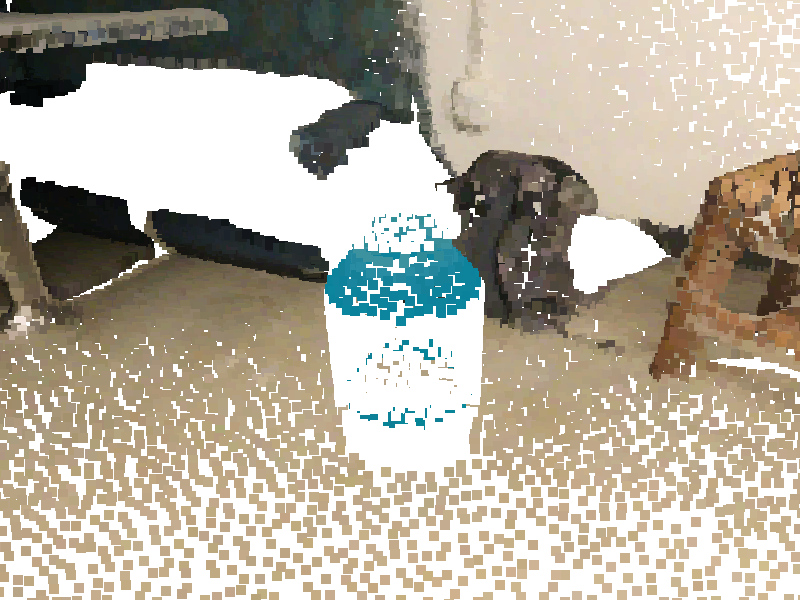} &
    \includegraphics[width=0.11\textwidth]{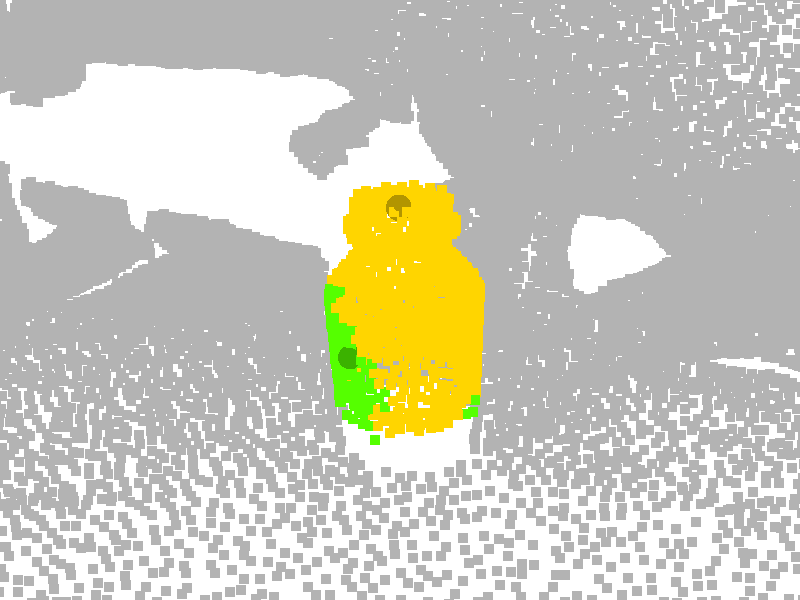} &
    \includegraphics[width=0.11\textwidth]{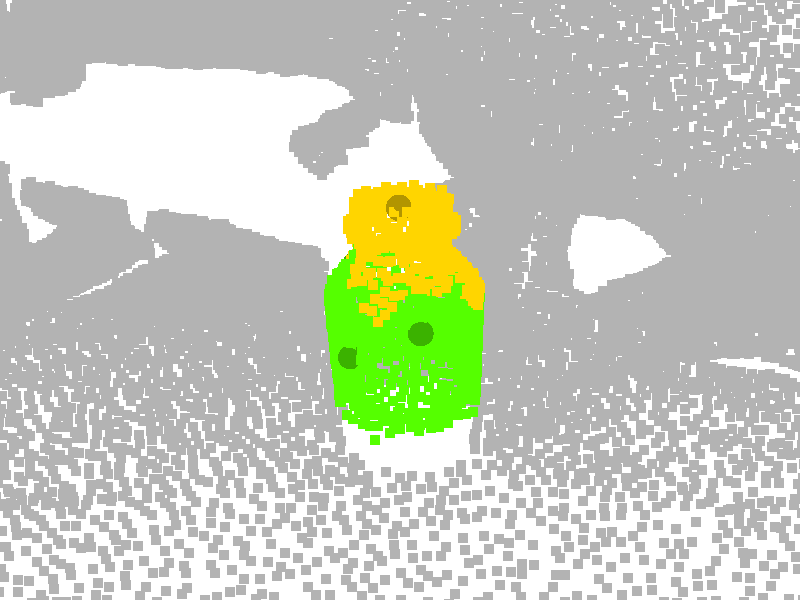} &
    \includegraphics[width=0.11\textwidth]{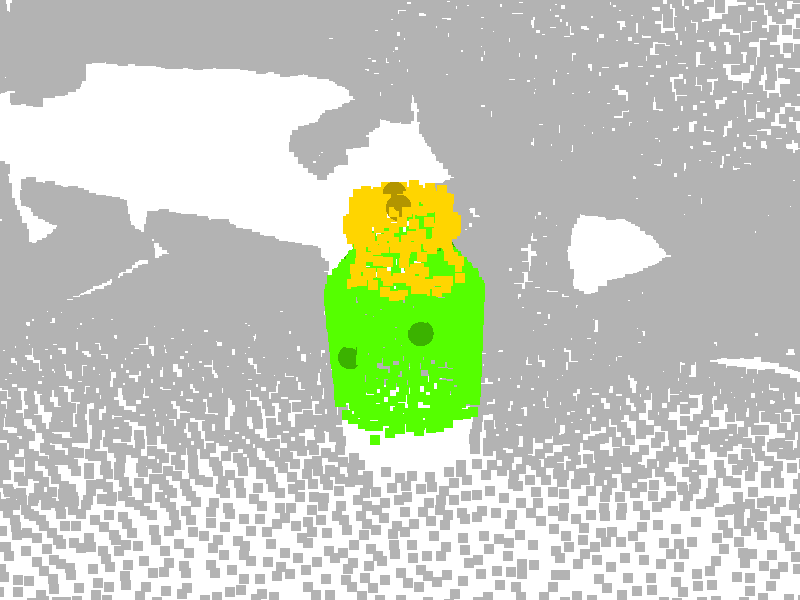} &
    \includegraphics[width=0.11\textwidth]{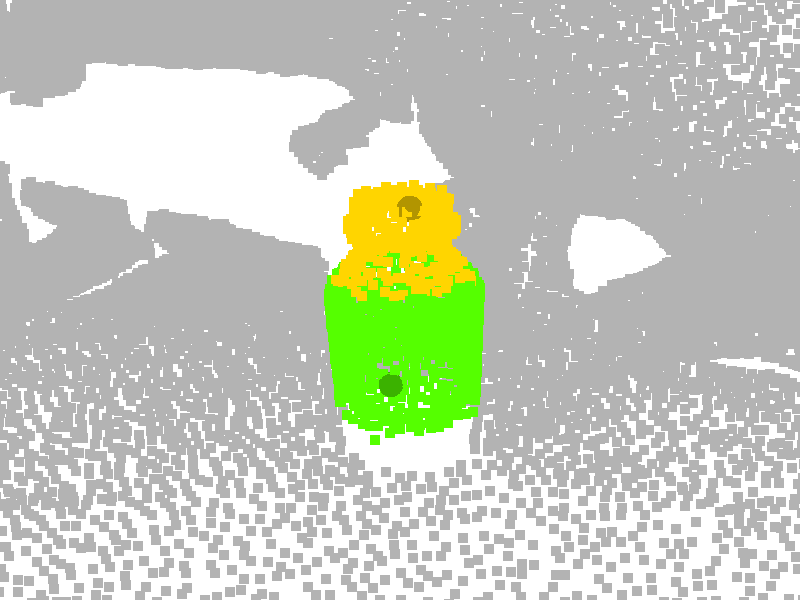} &
    \includegraphics[width=0.11\textwidth]{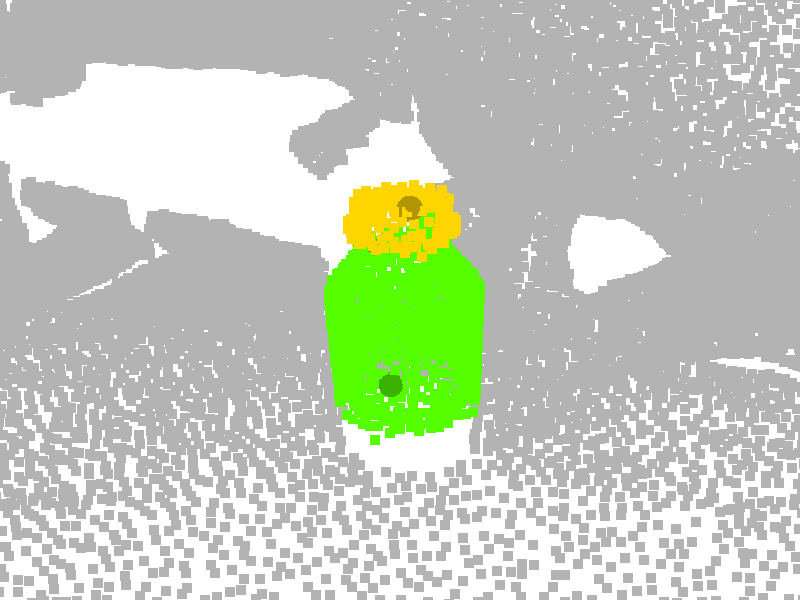} &
    \includegraphics[width=0.11\textwidth]{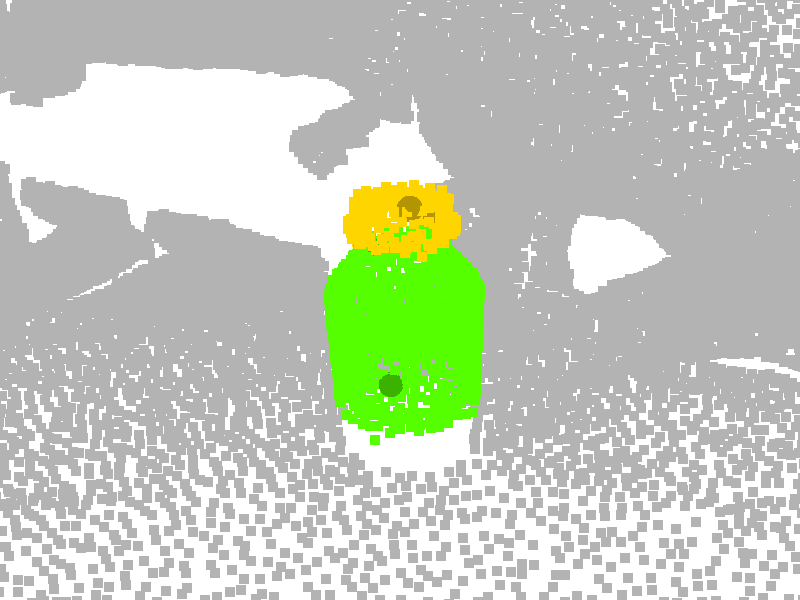} &
    \includegraphics[width=0.11\textwidth]{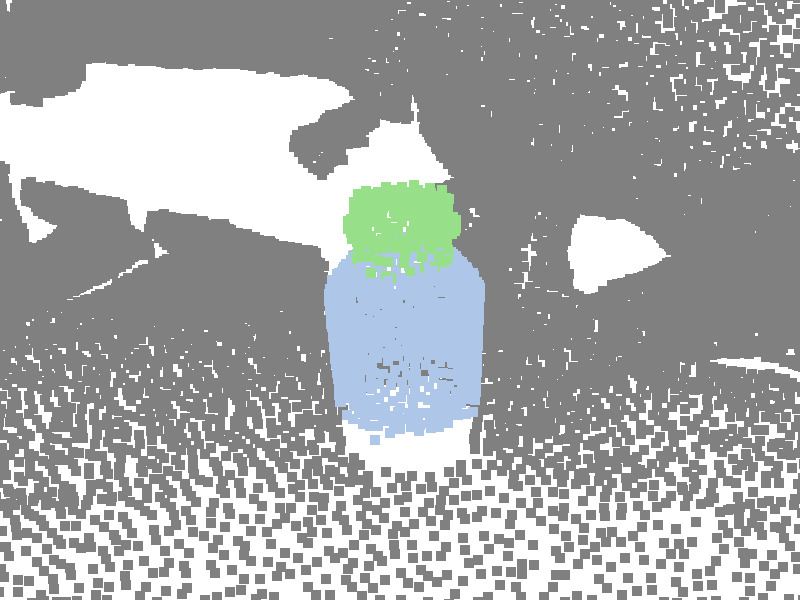} \\
    & & {\scriptsize $\overline{\text{IoU}}@2=18.8$} & {\scriptsize $\overline{\text{IoU}}@4=66.1$} & {\scriptsize $\overline{\text{IoU}}@6=79.8$} & {\scriptsize $\overline{\text{IoU}}@2=69.8$} & {\scriptsize $\overline{\text{IoU}}@3=89.0$} & {\scriptsize $\overline{\text{IoU}}@4=91.6$} & \\

    \multirow{3}{*}{\rotatebox{90}{MultiScan}} &
    \includegraphics[width=0.11\textwidth]{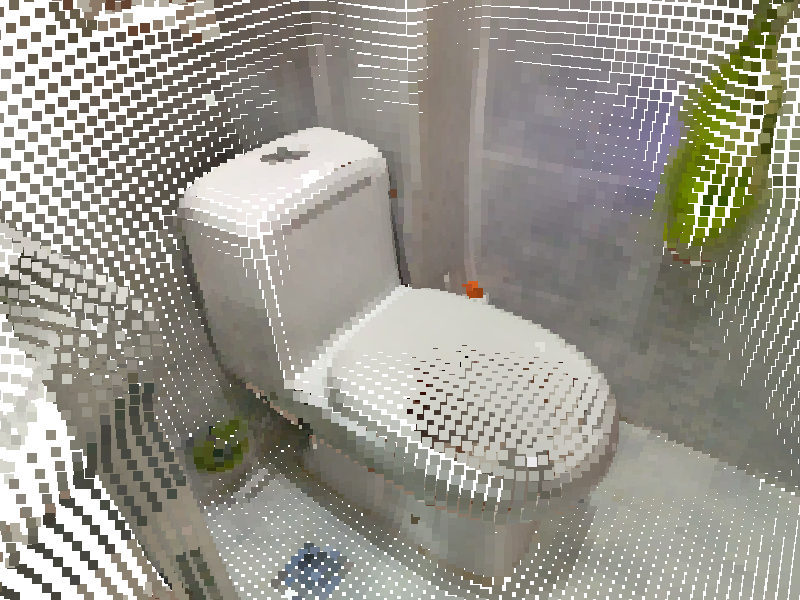} &
    \includegraphics[width=0.11\textwidth]{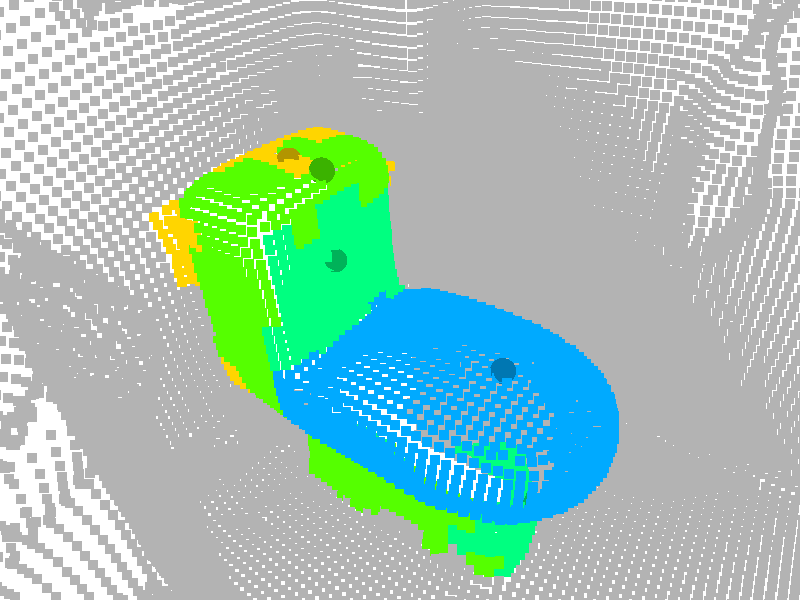} &
    \includegraphics[width=0.11\textwidth]{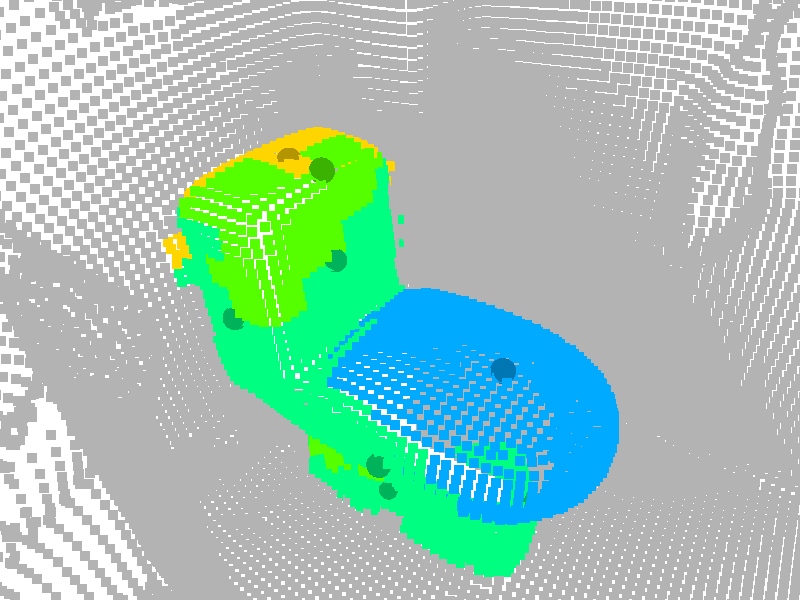} &
    \includegraphics[width=0.11\textwidth]{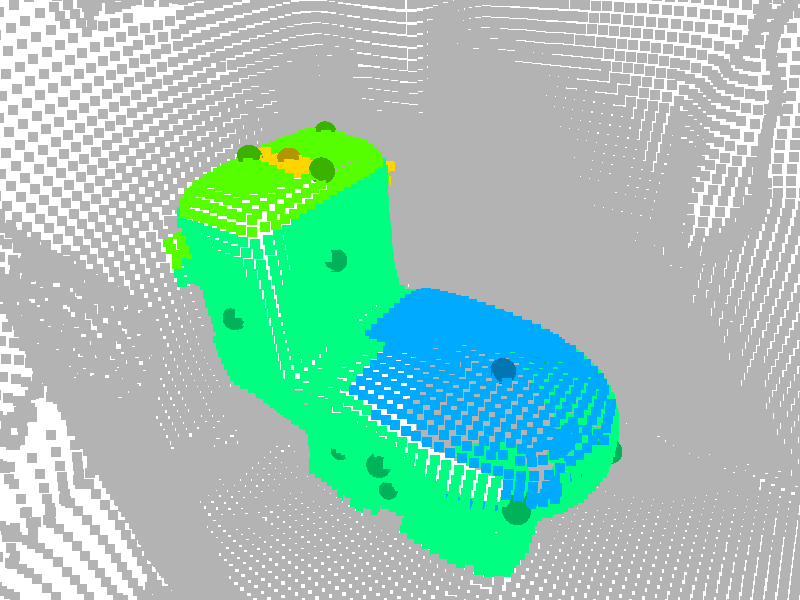} &
    \includegraphics[width=0.11\textwidth]{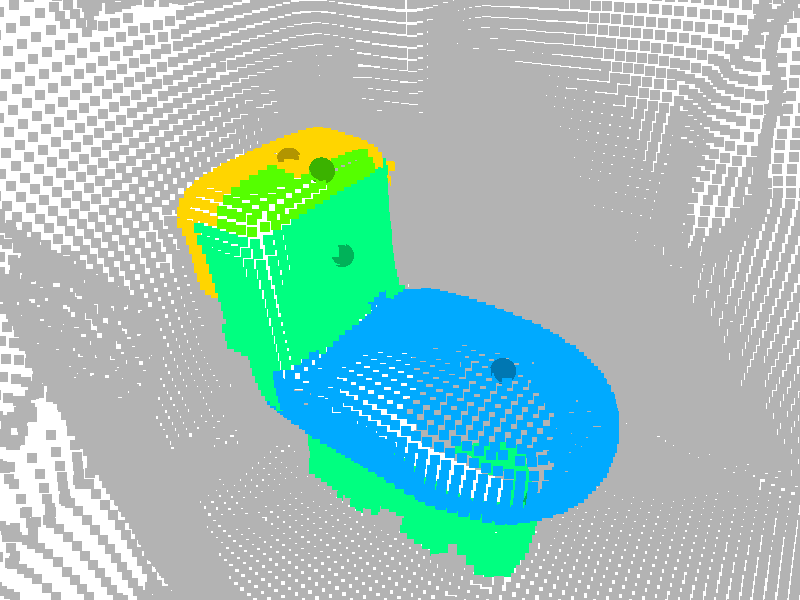} &
    \includegraphics[width=0.11\textwidth]{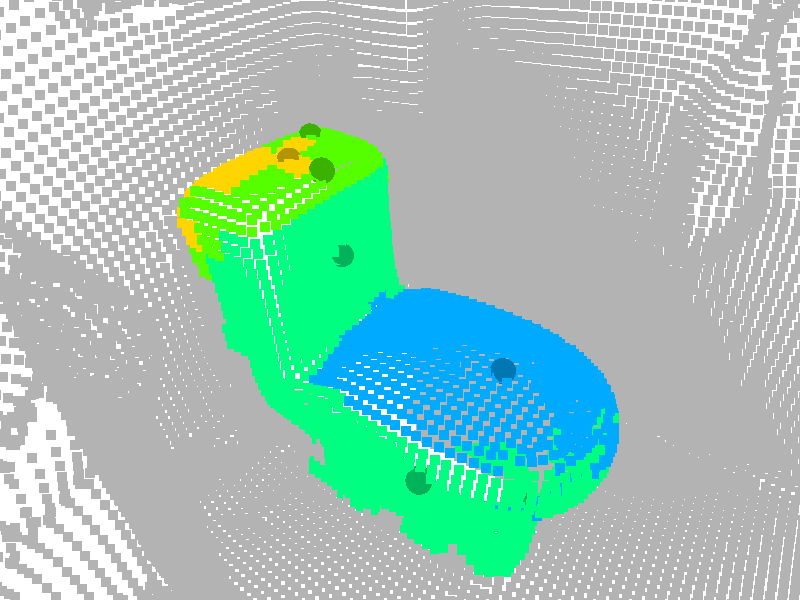} &
    \includegraphics[width=0.11\textwidth]{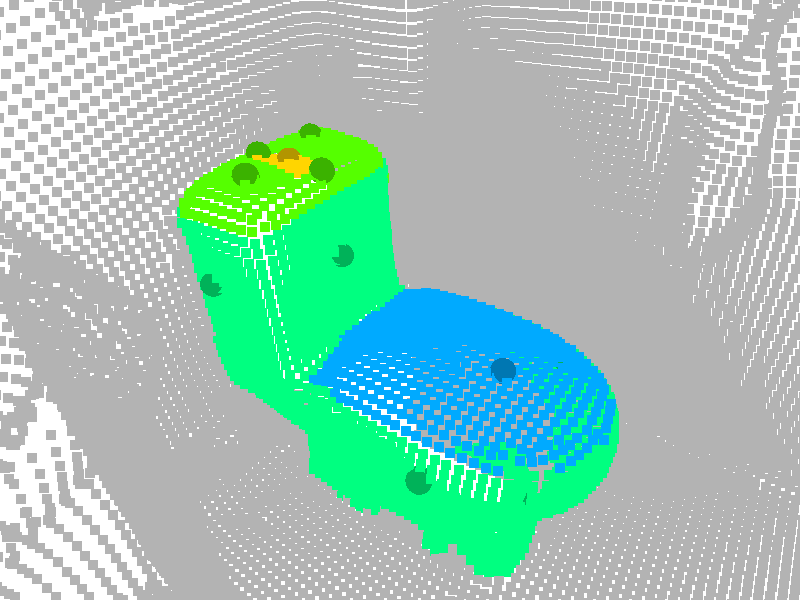} &
    \includegraphics[width=0.11\textwidth]{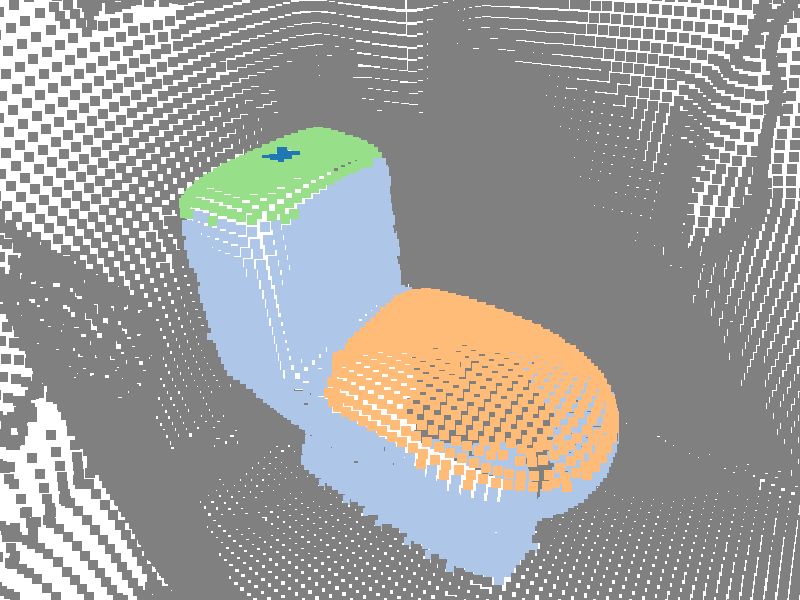} \\
    & & {\scriptsize $\overline{\text{IoU}}@5=23.7$} & {\scriptsize $\overline{\text{IoU}}@8=33.5$} & {\scriptsize $\overline{\text{IoU}}@14=58.7$} & {\scriptsize $\overline{\text{IoU}}@5=30.2$} & {\scriptsize $\overline{\text{IoU}}@7=48.3$} & {\scriptsize $\overline{\text{IoU}}@11=66.7$} & \\

    & \includegraphics[width=0.11\textwidth]{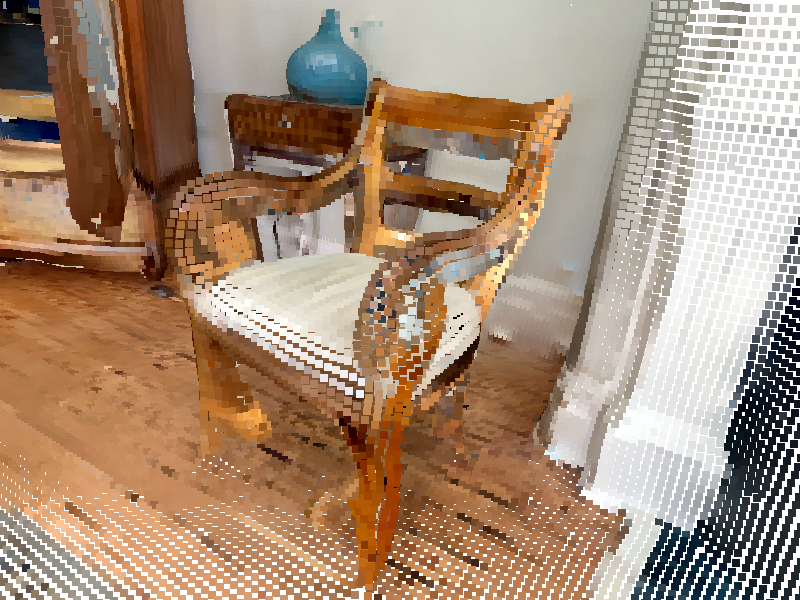} &
    \includegraphics[width=0.11\textwidth]{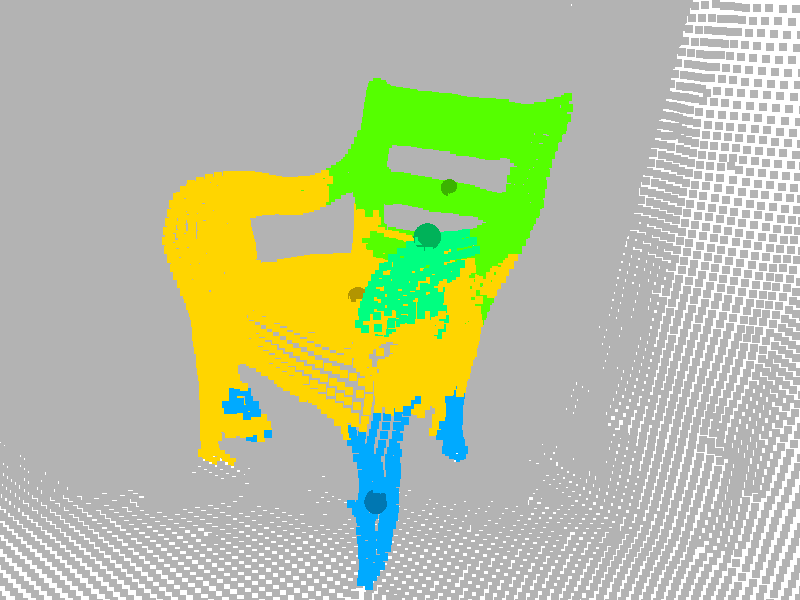} &
    \includegraphics[width=0.11\textwidth]{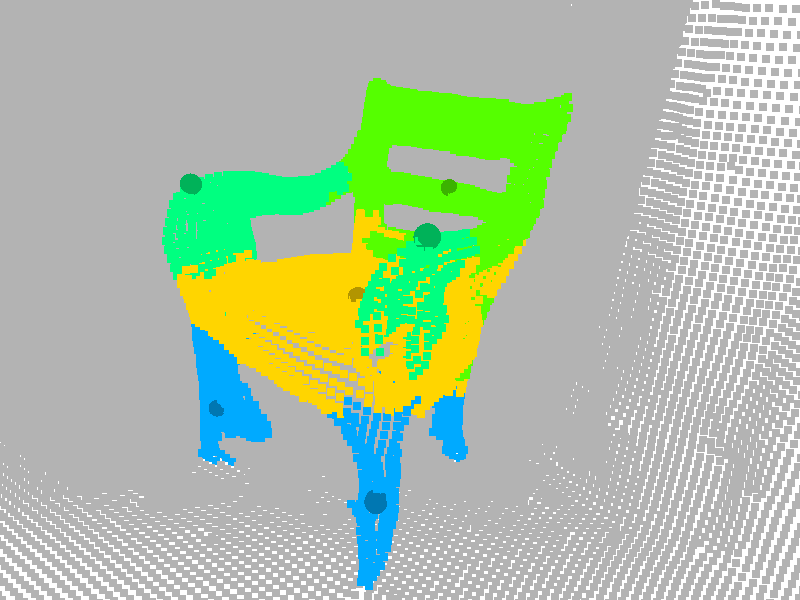} &
    \includegraphics[width=0.11\textwidth]{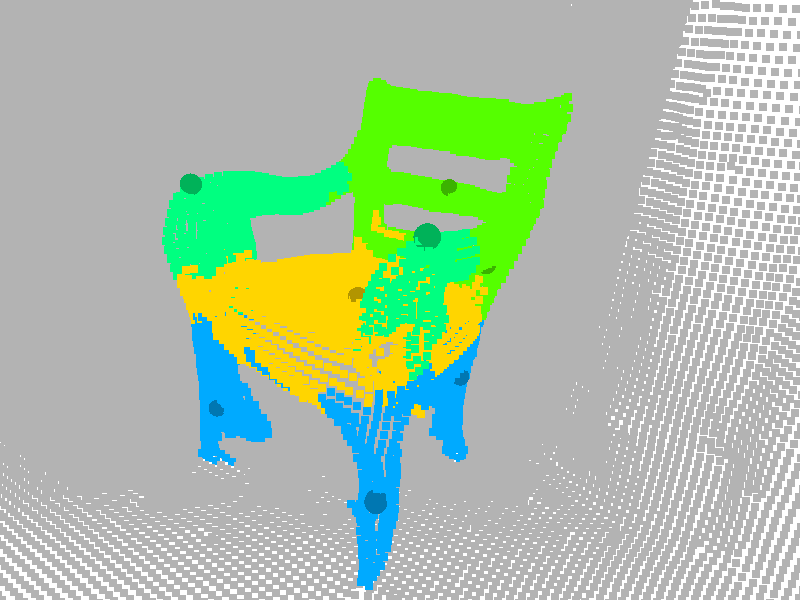} &
    \includegraphics[width=0.11\textwidth]{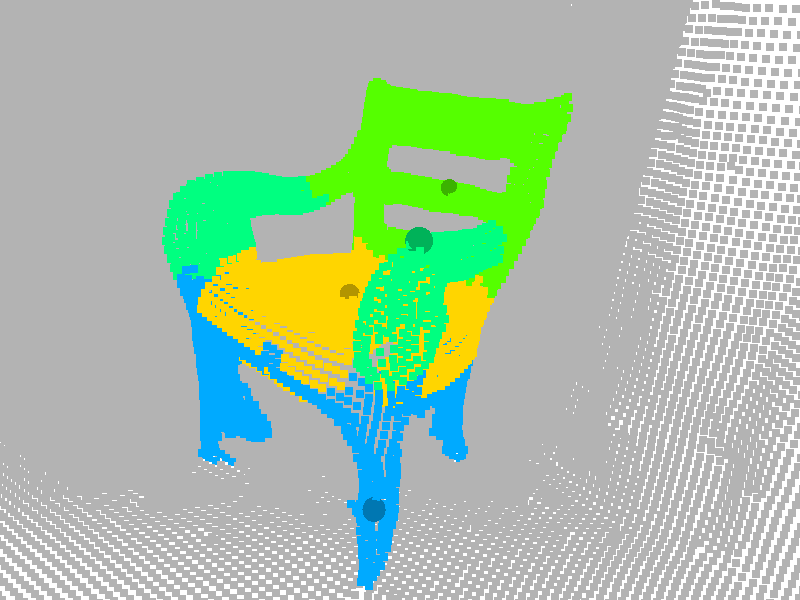} &
    \includegraphics[width=0.11\textwidth]{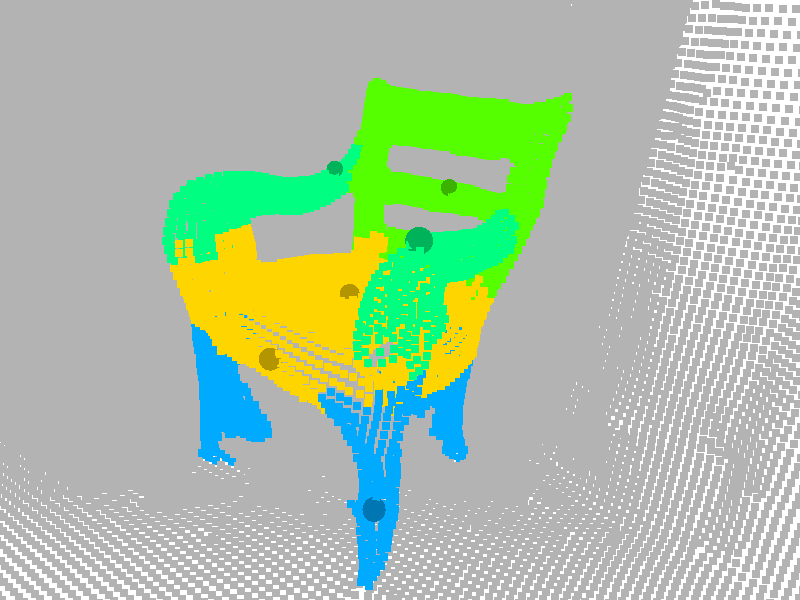} &
    \includegraphics[width=0.11\textwidth]{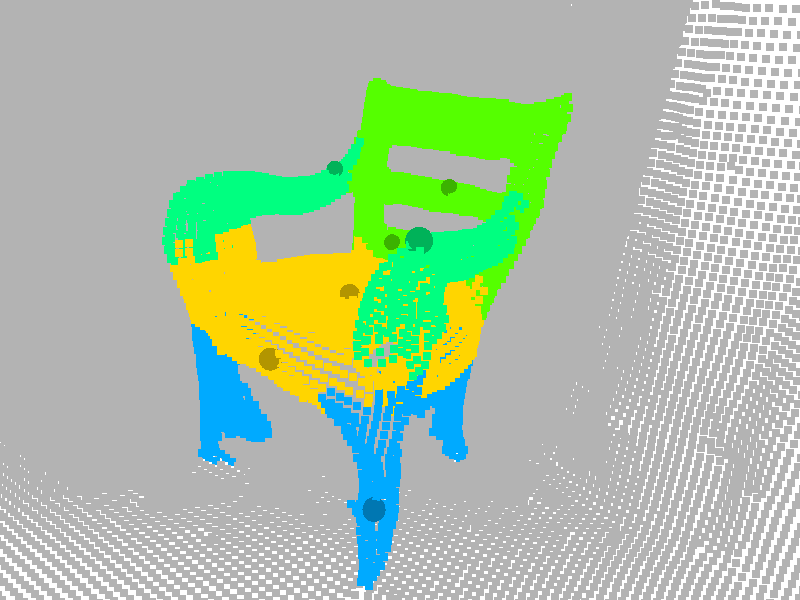} &
    \includegraphics[width=0.11\textwidth]{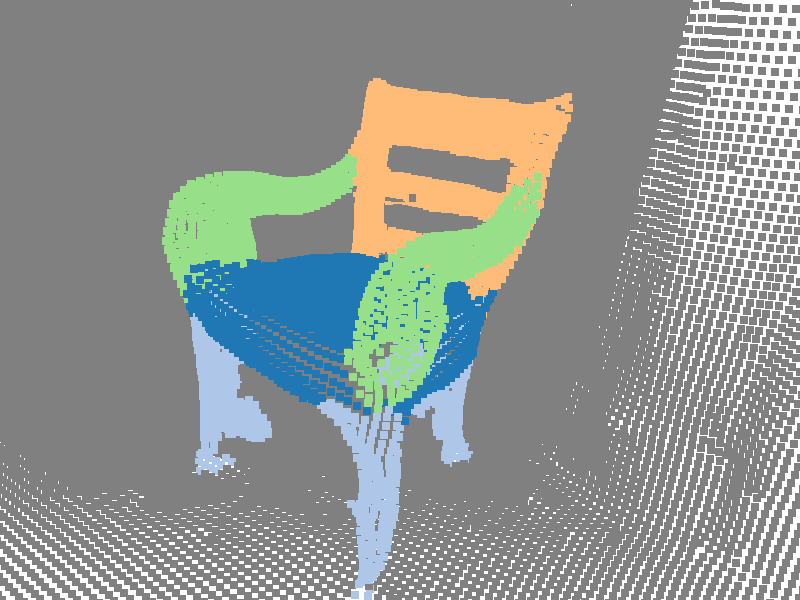} \\
    & & {\scriptsize $\overline{\text{IoU}}@4=39.1$} & {\scriptsize $\overline{\text{IoU}}@6=59.5$} & {\scriptsize $\overline{\text{IoU}}@8=63.4$} & {\scriptsize $\overline{\text{IoU}}@4=69.6$} & {\scriptsize $\overline{\text{IoU}}@5=71.9$} & {\scriptsize $\overline{\text{IoU}}@7=73.2$} & \\

    & \multicolumn{1}{c}{3DScene} 
    & \multicolumn{3}{c}{AGILE3D} 
    & \multicolumn{3}{c}{PinPoint3D (ours)} 
    & GroundTruth \\
\end{tabular}
}

\caption{\textbf{Qualitative comparison on interactive part segmentation.}}
\label{fig:qualitative-I}
\end{figure*}

\begin{figure*}[t]
\centering
\small
\setlength{\tabcolsep}{2pt} %
\renewcommand{\arraystretch}{0.9} %

\makebox[\textwidth][c]{%
\begin{tabular}{c:ccc}
    
    \includegraphics[width=0.22\textwidth]{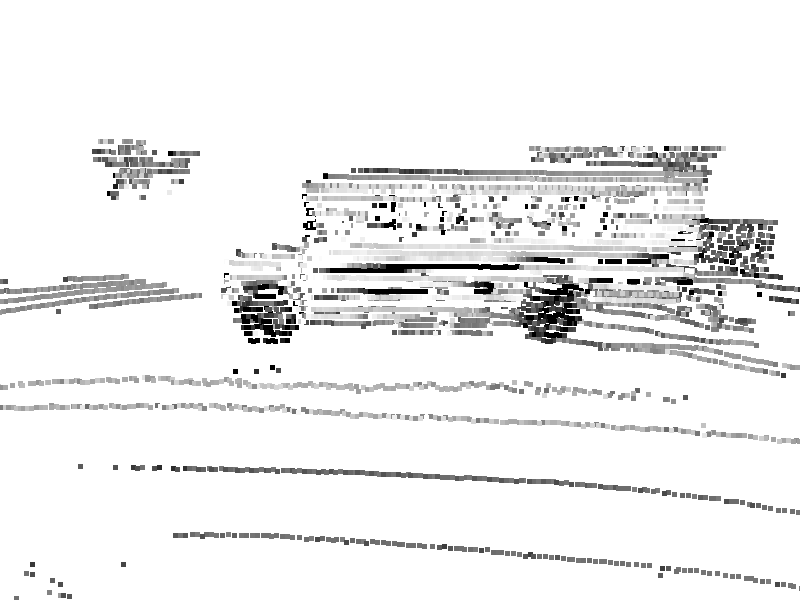} &
\includegraphics[width=0.22\textwidth]{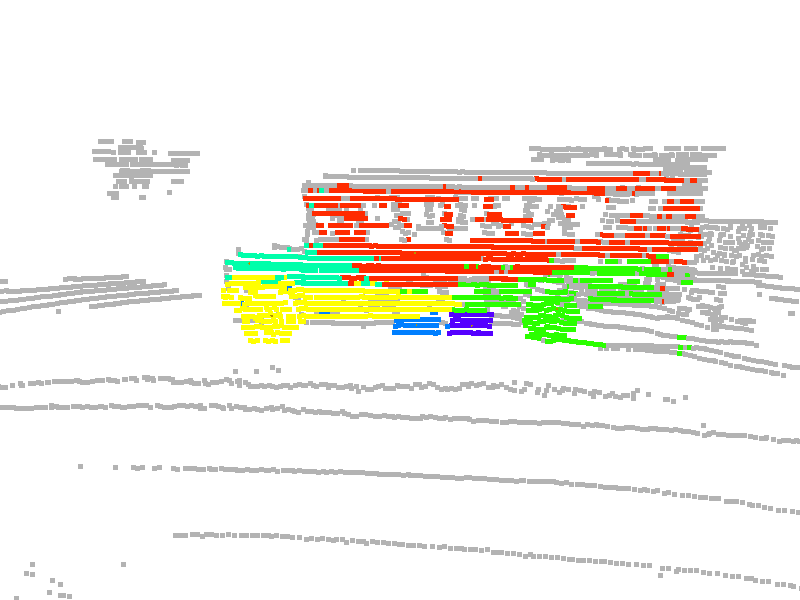} &
\includegraphics[width=0.22\textwidth]{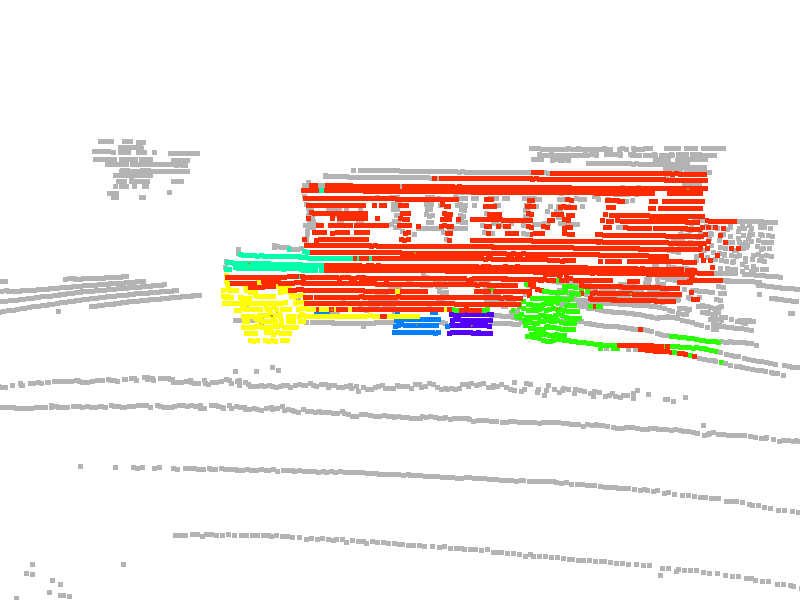} &
\includegraphics[width=0.22\textwidth]{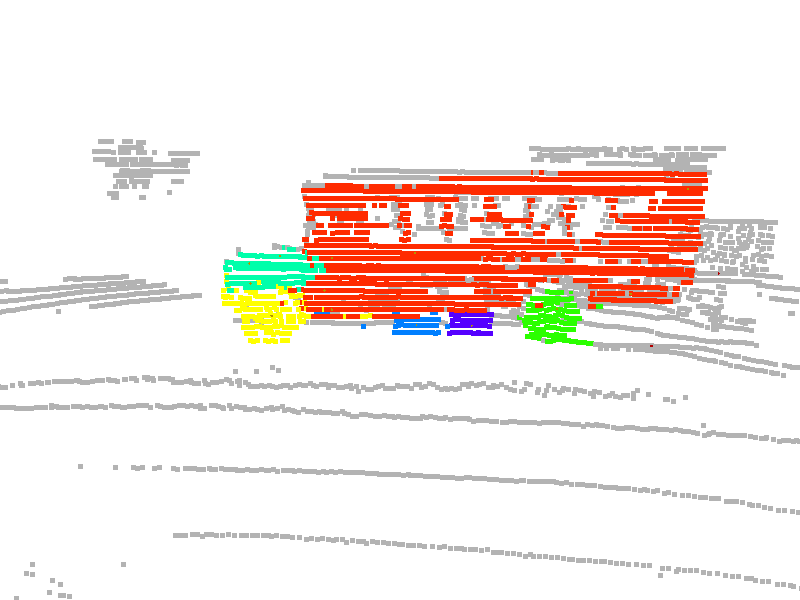}\\
KITTI & 1 click/part & 1.5 click/part &  2.5 click/part 

\end{tabular}
}

\caption{\textbf{Cross-dataset generalization test on KITTI Odometry Dataset}}
\label{fig:qualitative-one-row}
\end{figure*}

\section{Experiment}
In this section, we present the experimental results for PinPoint3D on the fine-grained 3D scene part segmentation task. Since no standardized evaluation protocol exists for this novel task, we adopt and extend the evaluation methodology of AGILE3D to suit part-level segmentation in 3D scenes. We evaluate our approach on both in-domain and cross-domain data for fine-grained part segmentation.


\noindent{\bf Datasets}  We trained PinPoint3D on our dataset PartScan, a synthesized dataset with part-level ground truth and real data with pseudo labels, as mentioned in section~\ref{sec:data-gen}. A dedicated test split is reserved to ensure no overlap with the training set, which provides human-verified part annotations for 12 categories derived from PartNet~\cite{mo2019partnet}. To assess the generalization ability of our model, we further adopt the MultiScan~\cite{mao2022multiscan} dataset. MultiScan is a large-scale RGB-D dataset with part-level annotations, containing 10,957 objects and 5,129 parts, making it well-suited for evaluating cross-domain generalization.

\noindent{\bf Evaluation metrics}
We compare the methods on (1) $NoC_{q}$\% $\downarrow$, the average number of clicks needed to reach $q$\% IoU, and (2) $IoU_{k}$ $\uparrow$, the average IoU achieved after $k$ user clicks per part (capped at 10). We also use $IoU@k$, which measures the IoU achieved after $k$ clicks per object.
 
For comparison with non-interactive segmentation methods, we additionally report weighted mIoU, which computes a weighted average of IoU across all parts according to the number of points in each part. This metric provides a more balanced evaluation of overall segmentation performance, especially when parts vary significantly in size.

\noindent{\bf Baseline}
We compare PinPoint3D against two baseline methods. PointSAM~\cite{zhou2025pointsam} is an interactive point-cloud segmentation model originally developed for object-level segmentation; the authors note that it can also be applied to part segmentation within scenes. The second baseline, AGILE3D, was originally designed for object segmentation but can produce part-level results through repeated user clicks.


\subsection{Evaluation on Scene-Part Segmentation}

Table~\ref{tab:eval-scene-part-seg} presents the comparative results on the scene part segmentation task. We adopt two testing strategies: the random-part strategy, where a randomly selected part of an object is segmented, and the all-part strategy, where all parts of an object are segmented. We report the $\text{NoC}_{q}\%$, $\text{IoU}_{k}$, and AP metrics for our model and the baselines. For PointSAM, the $\text{NoC}_{q}$ metric is not reported, as it often fails to reach the required IoU threshold on our test set.

In random-part segmentation tasks, our model achieves higher final IoU and requires fewer user clicks to reach a given IoU threshold. This advantage becomes even more pronounced in the all-part segmentation setting. This suggests PinPoint3D effectively captures the structural relationships among object parts, handling complex segmentation scenarios more efficiently, as further illustrated by representative annotation results in Fig.~\ref{fig:qualitative-I}. On the MultiScan dataset, however, our performance is nearly on par with AGILE3D. We attribute this to the relatively low complexity of the ground-truth annotations in MultiScan: while it provides rich object-level labels, the number of fine-grained part annotations is limited. Consequently, the benefits of our hierarchical design are less pronounced. We also conducted a cross-dataset generalization test on the KITTI Odometry dataset, with an example result shown in Fig.~\ref{fig:qualitative-one-row}.

\begin{table}[t]
\small
\centering
\caption{Comparison of Methods on Object Segmentation Ability}
\label{tab:eval-object-seg}
\renewcommand{\arraystretch}{1.2} %
\setlength{\tabcolsep}{8pt}       %
\resizebox{0.98\linewidth}{!}{%
\begin{tabular}{llccc}
\hline
Method & Test dataset & $\text{IoU}_{1}$  & $\text{IoU}_{3}$  & $\text{IoU}_{5}$  \\
\hline
AGILE3D & \multirow{2}{*}{PartScan} & 83.64 & 96.87 & 97.69  \\
\bf{PinPoint3D (ours)}  &                            & \textbf{86.7} & \textbf{97.0} & \textbf{98.0 }\\
\hline
AGILE3D & \multirow{2}{*}{Multiscan}    & \textbf{58.46} & \textbf{75.04} & \textbf{81.02}  \\
\bf{PinPoint3D (ours)}  &                            & 57.1 & 72.3 & 78.6  \\
\hline
\end{tabular}}
\end{table}

\subsection{Evaluation on Object Segmentation}

In this part, we evaluate the object segmentation capability of our model. Since our method is an extension of AGILE3D, we compare it with the original baseline. The result is presented in Table~\ref{tab:eval-object-seg}. The results show that our model achieves improved object segmentation performance on the synthetic dataset, while on the MultiScan generalization test set, the difference compared with AGILE3D is marginal. This indicates that our approach preserves the object-level segmentation ability of AGILE3D, while further enabling multi-granularity segmentation capability.


\begin{table*}[t!]
\centering
\small
\caption{Ablations on architecture and training strategy (evaluated on PartScan).}
\label{tab:ablate-arch-train}
\setlength{\tabcolsep}{6pt}
\renewcommand{\arraystretch}{1.15}
\begin{tabular*}{0.85\linewidth}{@{\extracolsep{\fill}} lcccccc}
\toprule
\textbf{Variant} & $\text{IoU}_{1}\uparrow$ & $\text{IoU}_{3}\uparrow$ & $\text{IoU}_{5}\uparrow$ & $\text{NoC}_{50}\downarrow$ & $\text{NoC}_{65}\downarrow$ & $\text{NoC}_{80}\downarrow$ \\
\midrule
Baseline (ours)            & 55.8 & \textbf{68.4} & \textbf{71.6} & \textbf{1.70} & \textbf{3.83} & \textbf{7.10} \\
\midrule
\multicolumn{7}{l}{\textit{Architecture}} \\
\quad No Part-Transformer & 43.7 & 60.2 & 65.0 & 2.54 & 5.46 & 8.60 \\
\quad No Adapter (Unfreezing Backbone)   & \textbf{56.3} & 67.5 & 70.4 & 1.73 & 3.97 & 7.40 \\
\midrule
\multicolumn{7}{l}{\textit{Training Strategy}} \\
\quad Single-Object        & 52.5 & 67.5 & 70.8 & 1.79 & 3.96 & 7.38 \\
\bottomrule
\end{tabular*}
\end{table*}


\begin{table*}[t!]
\centering
\small
\caption{Ablations on training data (evaluated on MultiScan).}
\label{tab:ablate-data}
\setlength{\tabcolsep}{3pt}
\renewcommand{\arraystretch}{1.15}
\begin{tabular*}{0.85\textwidth}{@{\extracolsep{\fill}} lcccccc}
\toprule
\textbf{Data Variant} & $\text{IoU}_{1}\uparrow$ & $\text{IoU}_{3}\uparrow$ & $\text{IoU}_{5}\uparrow$ & $\text{NoC}_{50}\downarrow$ & $\text{NoC}_{65}\downarrow$ & $\text{NoC}_{80}\downarrow$ \\
\midrule
PartScan (ours) & \textbf{44.4} & \textbf{62.7} & \textbf{68.1} & \textbf{2.37} & \textbf{4.86} & 8.06 \\
PartNet (in-scene)           & 43.3 & 62.4 & 67.8 & 2.50 & 4.91 & 8.10 \\
ScanNet-PartField            & 43.1 & 61.6 & 67.4 & 2.65 & 5.10 & \textbf{8.05} \\
\bottomrule
\end{tabular*}
\end{table*}


\begin{table}[t!]
\centering
\small
\begin{threeparttable}
\caption{Effect of adapter/backbone strategy on object segmentation.}
\label{tab:tam_training_protocol}
\begin{tabular}{lcccc}
\toprule
Protocol & Obj IoU@1 & Obj IoU@3 & $\text{AP}_{25\%}\uparrow$ & $\text{AP}_{50\%}\uparrow$ \\
\midrule
A+F   & \textbf{86.7} & \textbf{97.0} & \textbf{90.6} & \textbf{89.9} \\
NA+UF & 58.9 & 92.2 & 88.5 & 85.6 \\
\bottomrule
\end{tabular}
\begin{tablenotes}
\footnotesize
\centering
\item A+F = Adapter + Frozen Backbone;\\ NA+UF = No Adapter + Unfrozen Backbone.
\end{tablenotes}
\end{threeparttable}
\end{table}


\begin{figure*}[!t]
\centering
\small
\setlength{\tabcolsep}{2pt}
\renewcommand{\arraystretch}{0.9}
\begin{tabular}{cccc}
    \includegraphics[width=0.24\textwidth]{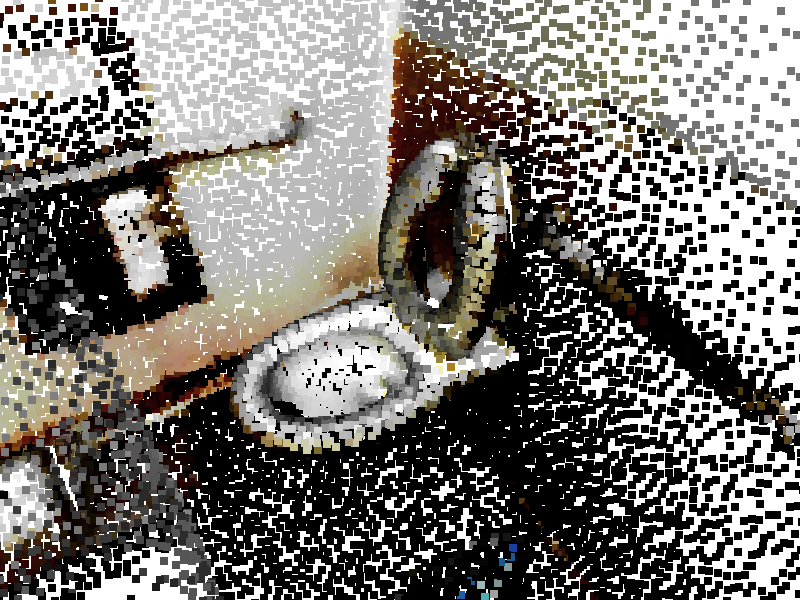} &
    \includegraphics[width=0.24\textwidth]{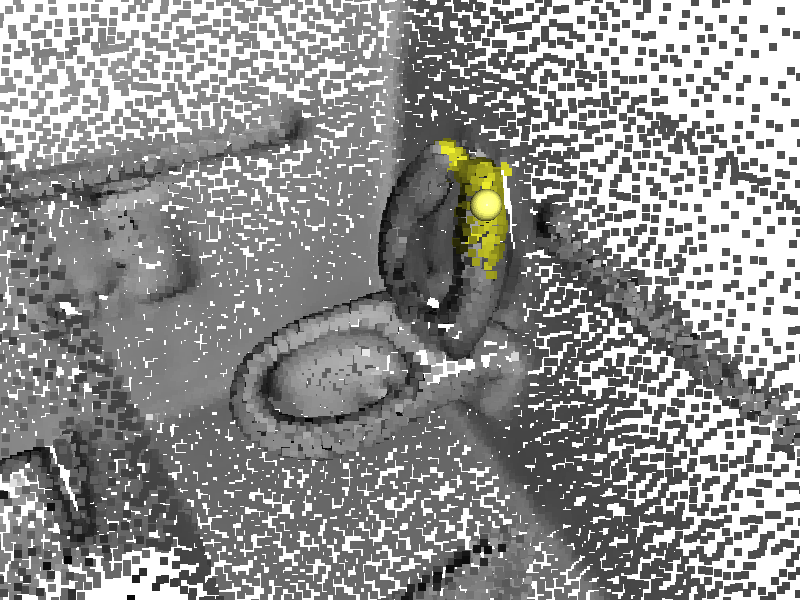} &
    \includegraphics[width=0.24\textwidth]{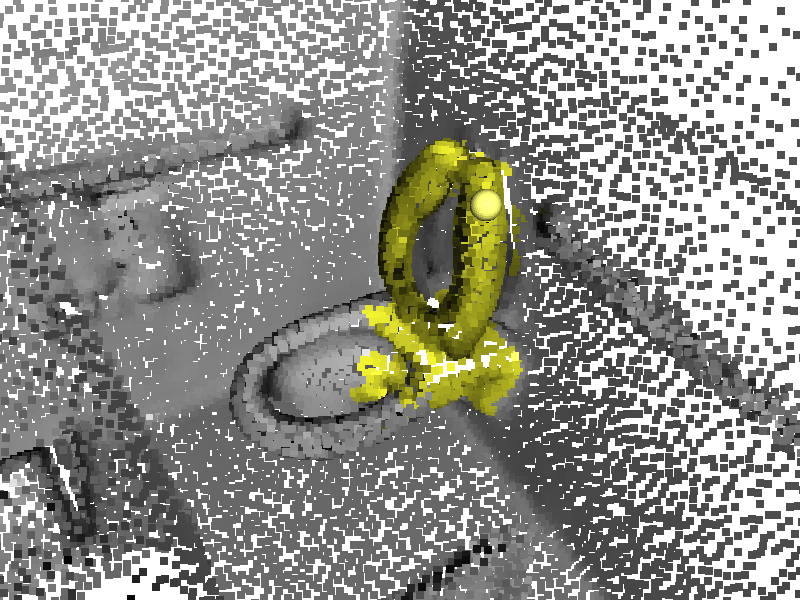} &
    \includegraphics[width=0.24\textwidth]
    {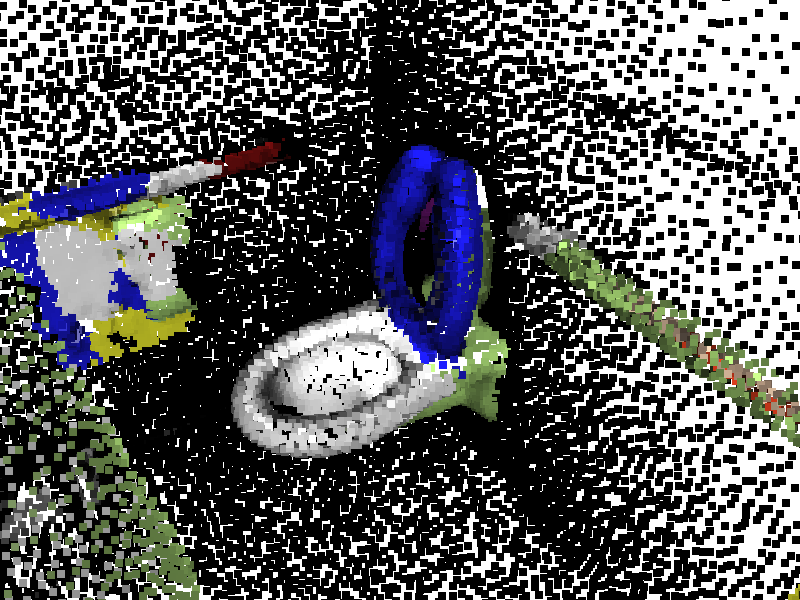}\\
    (a) ScanNet & (b) Single-Object Training & (c) Multi-Object Training & (d) Pseudo parts on ScanNet
\end{tabular}
\caption{\textbf{Qualitative comparison of training protocols.} 
Single-object training often suffers from excessive competition and coupling within self-attention, causing each click to cover only a small region and producing fragmented masks. 
In contrast, multi-object training alleviates this over-coupling,  cover a larger extent with each click, resulting in more complete part coverage.}
\label{fig:qualitative-single-row-multi}
\end{figure*}

\section{Ablation study}
\subsection{Architecture design}
We first examine two key architectural design choices: the adapter in the feature encoder and a dedicated part transformer branch.

For the adapter, we compare direct fine-tuning of the backbone with using a frozen backbone plus lightweight adapters. As shown in Tab.~\ref{tab:ablate-arch-train}, direct fine-tuning yields only marginal part-level gains (IoU@1) while destabilizing object-level representations (Tab.~\ref{tab:tam_training_protocol}). In contrast, adapters preserve the semantic feature space and maintain strong object-level accuracy, while still providing significant improvements in part segmentation.

We also evaluate the necessity of the instance-part transformer decoder. Previous methods on Interactive Instance Segmentation can be adapted to part segmentation by adding more point prompts for filtering.
Thus, we remove the part transformer and train a unified decoder, directly utilizing the mask module to predict part masks from the output of the object-level decoder. The \textbf{Architecture} block of Table.~\ref{tab:ablate-arch-train} indicates that the model's part segmentation accuracy drops sharply without the part transformer, since partitioning relies only on coarse object-level features. Including the part transformer enables a progressive refinement of representations from object-level to part-level, yielding improvements in fine-grained part segmentation performance.

\subsection{Training Strategy}
In the previous part segmentation tasks, training is typically based on a single object setting, which means only one targeted object is partitioned at a time. Such a protocol naturally ensures that part predictions are strictly constrained within the object. In contrast, scene-level part segmentation introduces multiple objects simultaneously, which raises another training protocol: whether to allow part learning across different objects.



{\bf Single- vs. Multi-Object Training.}
We compare two training protocols for the part decoder: single-object, where only one object's parts are clicked per iteration, and multi-object, where clicks from different objects appear jointly. Since the decoder still receives a single target ID, the latter forces it to interpret dispersed clicks as a ``composite object.'' As shown in the Training Strategy block of Tab.~\ref{tab:ablate-arch-train}, multi-object training consistently yields higher part IoU.

We attribute this improvement to three factors: (i) multi-object sampling introduces more dispersed and heterogeneous click distributions, forcing the model to rely on local evidence rather than assuming all clicks belong to a single connected region; (ii) the frozen object decoder tends to produce over-merged masks, which compels the part decoder to learn corrective splitting, transferring to single-object evaluation as more conservative mask growth and sharper boundaries; and (iii) queries sampled from more distant and heterogeneous regions thereby reduce excessive coupling (see Fig.~\ref{fig:qualitative-single-row-multi}), and under the combined influence of loss constraints and residual coordination in the shared scene, undergo soft competition that makes them act as specialized local experts---achieving more complete coverage while maintaining boundaries as clean as possible.

\subsection{Training Dataset Composition}
We study the impact of training dataset composition on segmentation performance. We compare training the model on: (i) PartNet-only, using synthetic objects with fine part annotations integrated into scenes; (ii) pseudo-label only, using real scanned scenes with pseudo-labeled parts generated by PartField; and (iii) a combined dataset PartScan containing both sources. All models are evaluated identically on the MultiScan validation set. Table~\ref{tab:ablate-data} shows: the combined PartScan data yields consistently better performance by uniting PartNet's diversity with ScanNet's realism.



\section{Limitations}

Our training dataset is primarily constructed from indoor scene scans, which introduces a generalization gap when applying the model to outdoor environments. Part of our dataset relies on pseudo-labels generated by existing algorithms rather than human-verified annotations, which may limit the achievable upper bound of segmentation accuracy. Furthermore, the limited availability of in-scene part-level annotations constrains the diversity of training data, which affects the generalization ability of our model to unseen categories and environments.






\bibliographystyle{plain}   

\clearpage

\twocolumn[{%
\centering
\vspace{2ex}
{\LARGE \textbf{APPENDIX}\par}
\vspace{3ex}
}]


\section*{ADDITIONAL RESULTS}

\subsection*{Quanlitative Results on KITTI-360}

\begin{figure*}[t!]
\centering
\setlength{\tabcolsep}{2pt} %
\renewcommand{\arraystretch}{0.9} %

\makebox[\textwidth][c]{
\begin{tabular}{c:ccc}
    \includegraphics[width=0.22\textwidth]{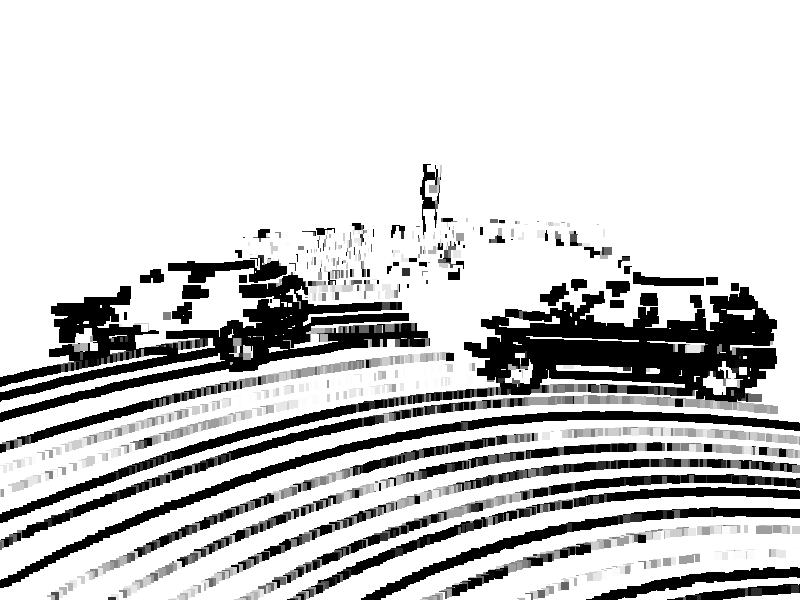} &
    \includegraphics[width=0.22\textwidth]{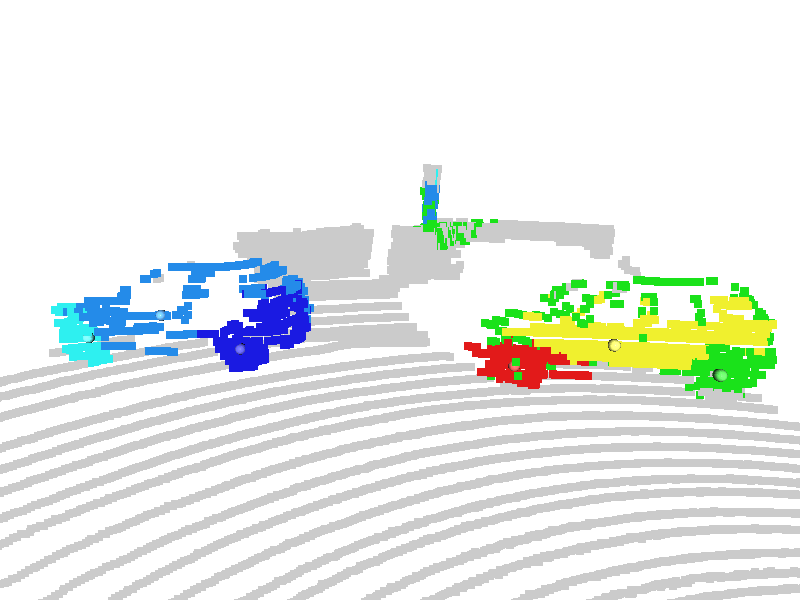} &
    \includegraphics[width=0.22\textwidth]{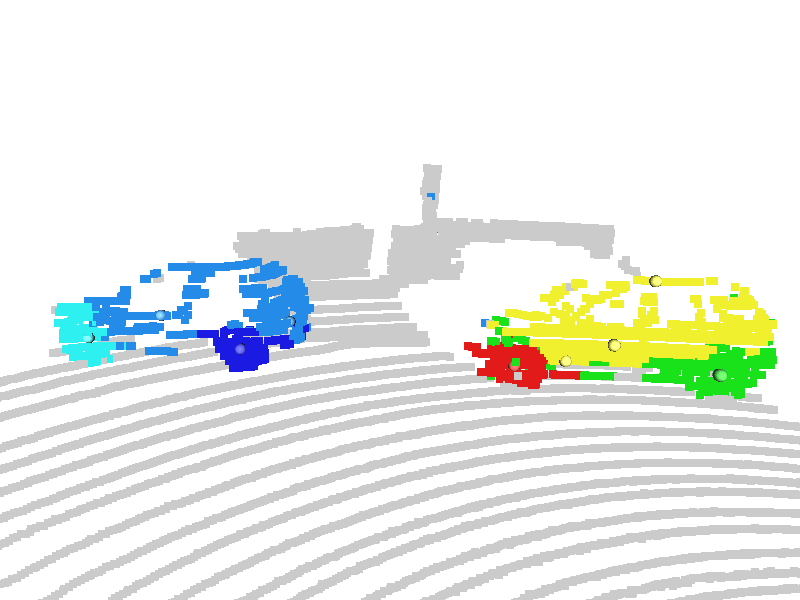} &
    \includegraphics[width=0.22\textwidth]{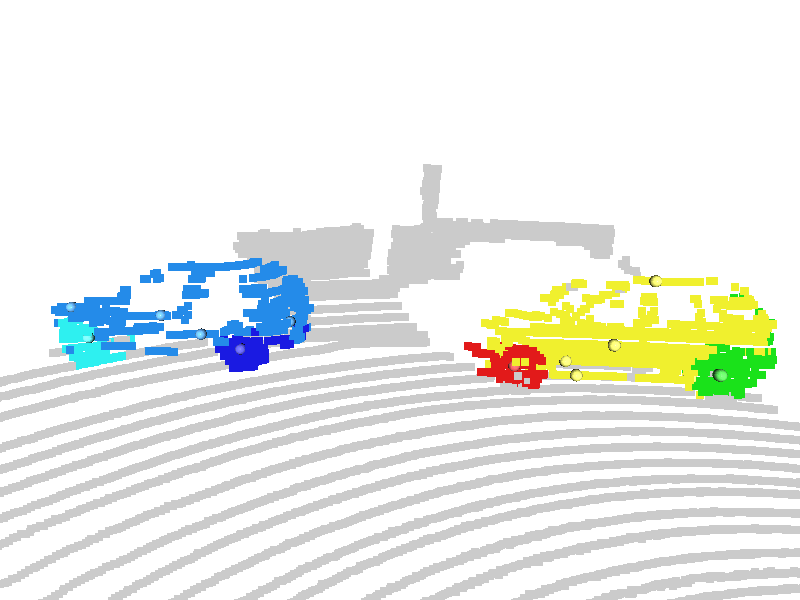} \\
    KITTI-360 & 1 click/part & 1.67 click/part & 2.33 click/part \\
    \includegraphics[width=0.22\textwidth]{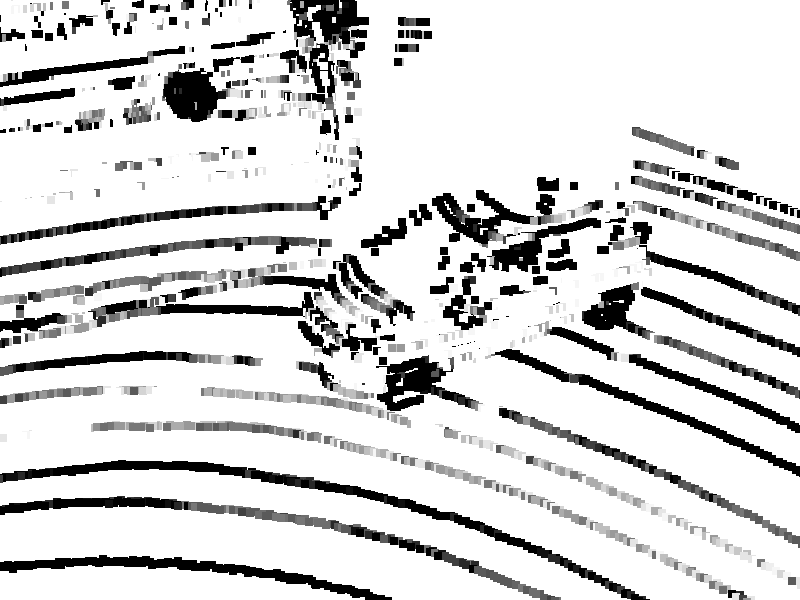} &
    \includegraphics[width=0.22\textwidth]{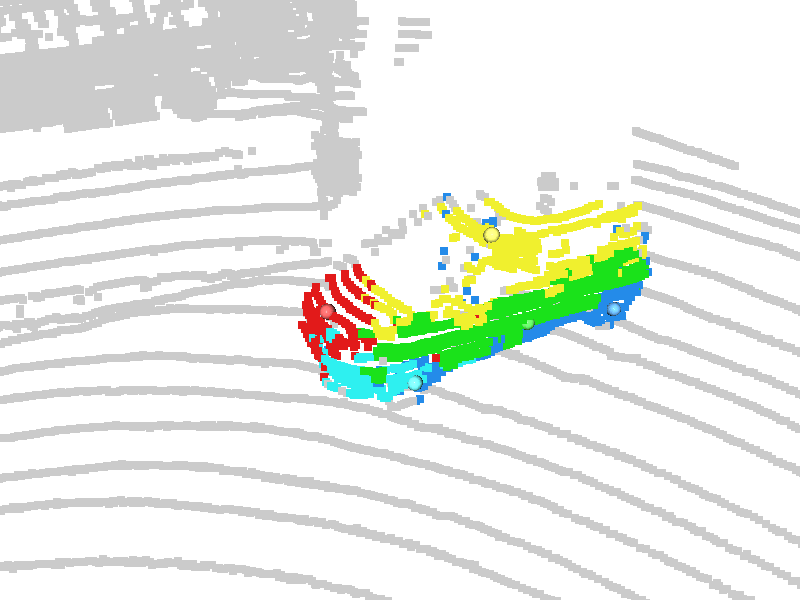} &
    \includegraphics[width=0.22\textwidth]{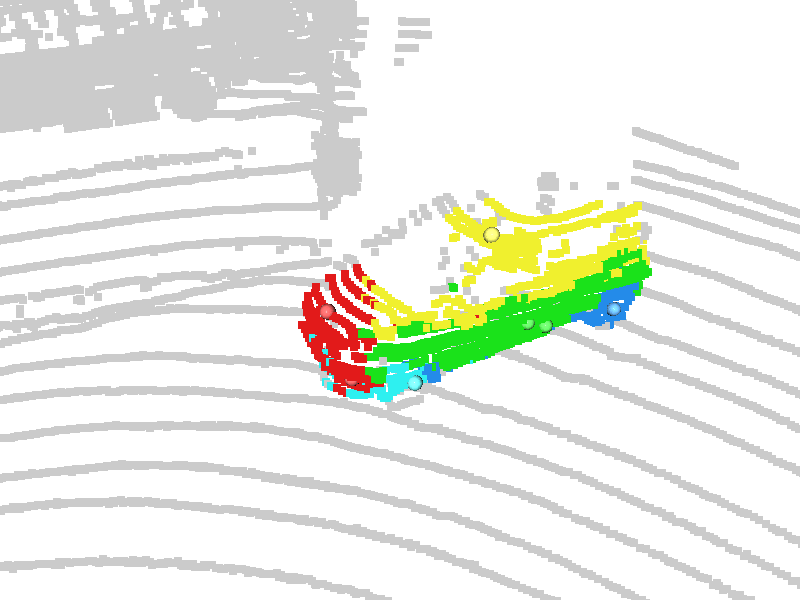} &
    \includegraphics[width=0.22\textwidth]{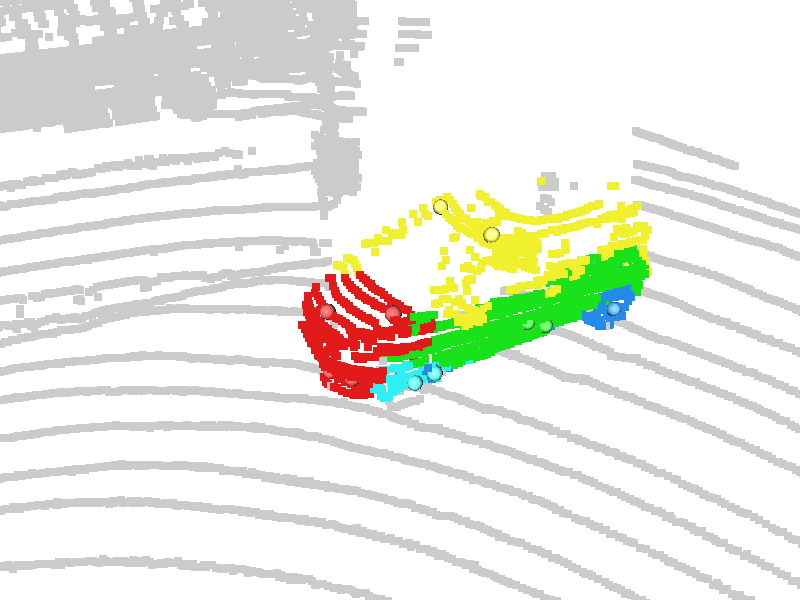} \\
    KITTI-360 & 1 click/part & 1.4 click/part & 2.1 click/part \\
\end{tabular}
}

\caption{\textbf{Qualitative results on interactive multi-object part segmentation on KITTI Odometry.}
We evaluate our model on outdoor driving scenes from KITTI. Results show progressive refinement of part masks as the average number of clicks per part increases, demonstrating effective segmentation performance in challenging real-world scenes.}

\label{fig:qualitative-two-rows}
\end{figure*}

In Fig.~\ref{fig:qualitative-two-rows}, we present additional qualitative results on the KITTI~\cite{kitti-360} driving scenes, which feature sparse long-range LiDAR, frequent occlusions, and large variations in viewpoint and scale. 
Starting from a single positive click on the target part, our model progressively refines both object- and part-level masks as additional clicks are provided. 
Despite the domain shift from indoor training data to outdoor driving environments, the model is able to yield relatively clean object boundaries on large textureless surfaces (e.g., car bodies), while assigning specialized queries to slender or articulated components (e.g., wheels, hood, side panels, roof rails). 
When the average number of clicks per part slightly exceeds two, the model can already recover most of the expected parts with reasonable accuracy. 
These results demonstrate that the proposed PinPoint framework generalizes well to challenging outdoor scenarios, requiring only minimal user effort.

\subsection*{Qualitative Results on PartScan}
\begin{figure*}[t]
\vspace*{5pt}
\centering
\small
\setlength{\tabcolsep}{1pt} %
\renewcommand{\arraystretch}{0.7} %

\makebox[\textwidth][c]{%
\begin{tabular}{cc|ccc|c}
    \multirow[c]{6}{*}[-8ex]{\rotatebox[origin=c]{90}{\textbf{PartScan}}} &
    \includegraphics[width=0.18\textwidth]{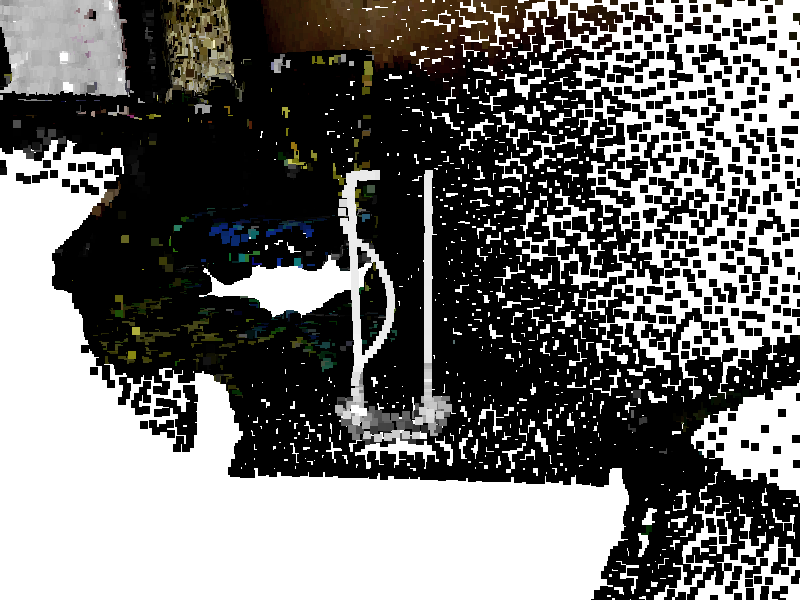} &
    \includegraphics[width=0.18\textwidth]{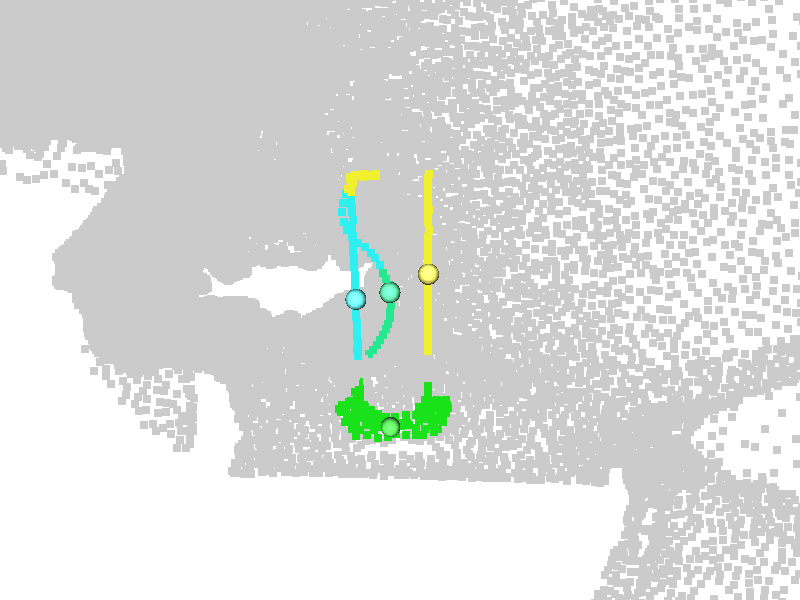} &
    \includegraphics[width=0.18\textwidth]{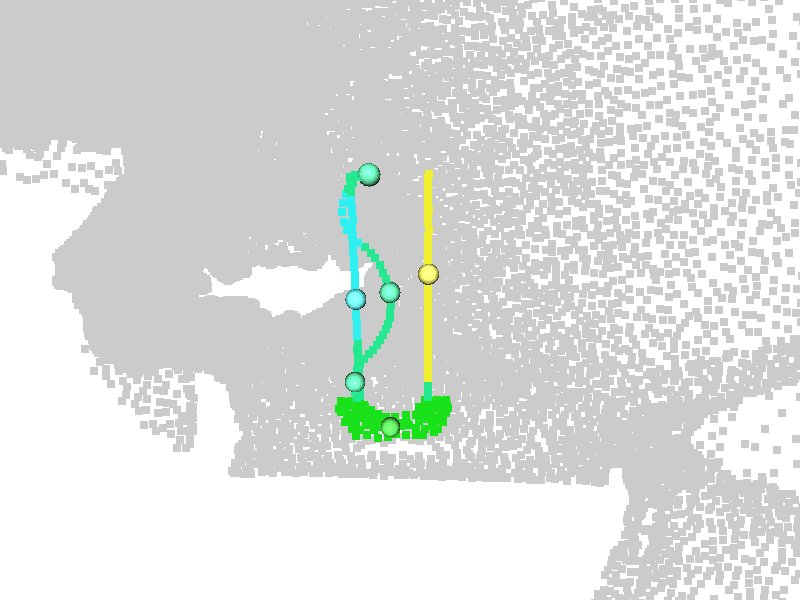} &
    \includegraphics[width=0.18\textwidth]{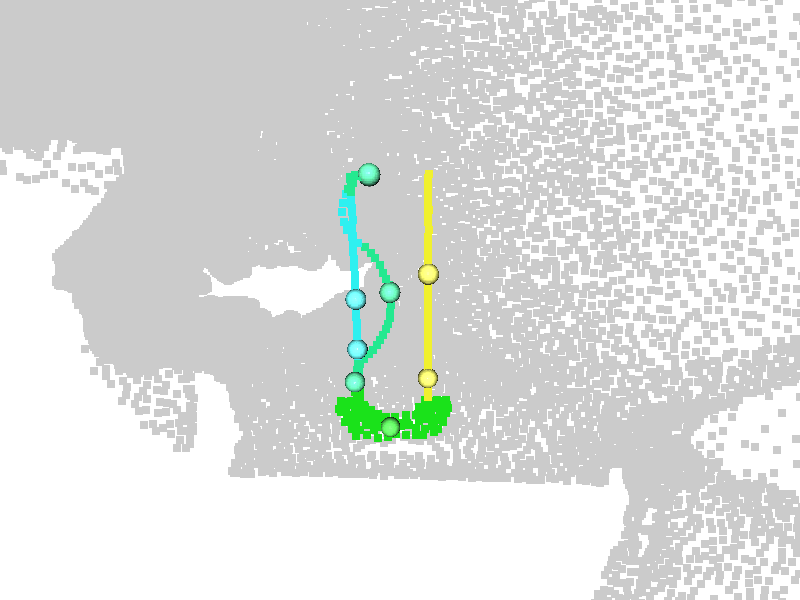} &
    \includegraphics[width=0.18\textwidth]{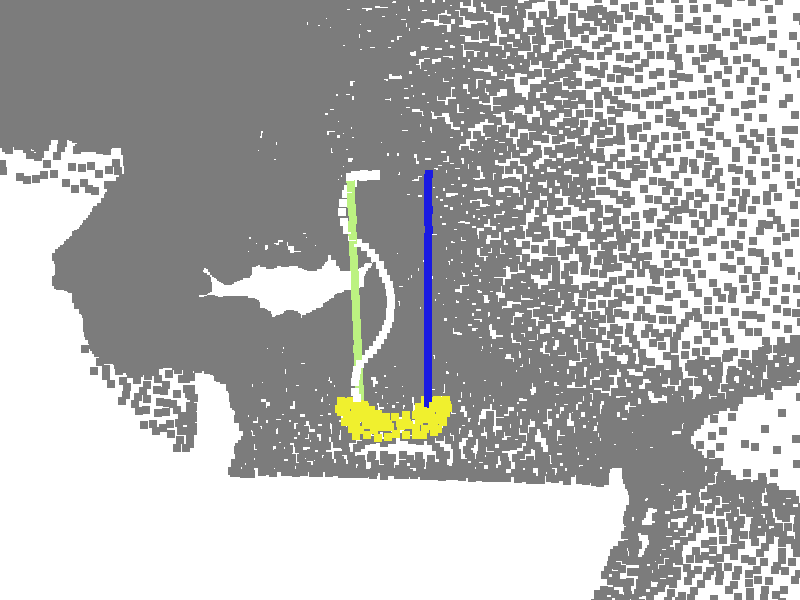} \\
    & & {\scriptsize $\overline{\text{IoU}}@4=55.7$} & {\scriptsize $\overline{\text{IoU}}@6=72.9$} & {\scriptsize $\overline{\text{IoU}}@8=79.5$} & \\

    & \includegraphics[width=0.18\textwidth]{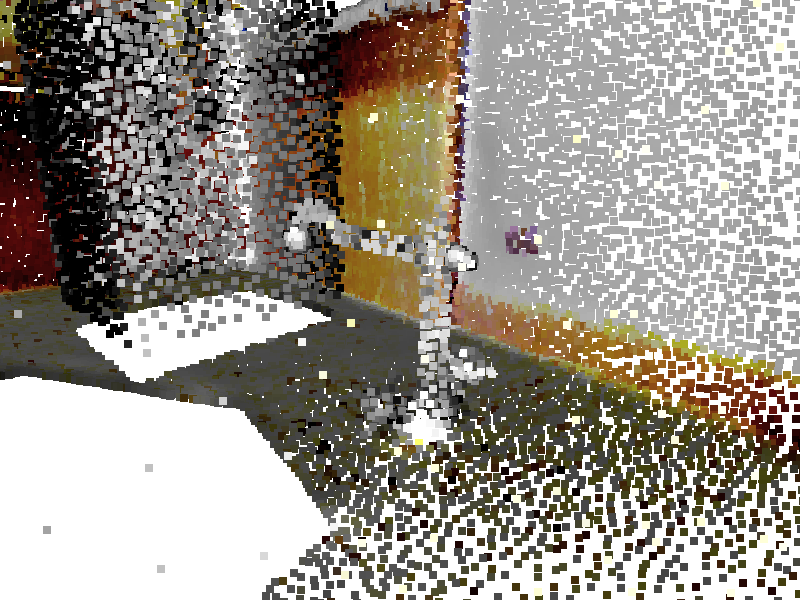} &
    \includegraphics[width=0.18\textwidth]{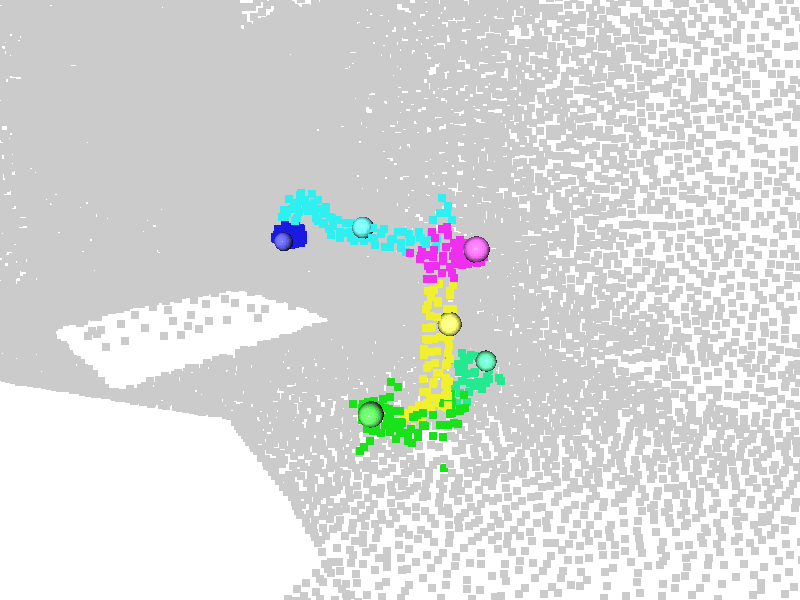} &
    \includegraphics[width=0.18\textwidth]{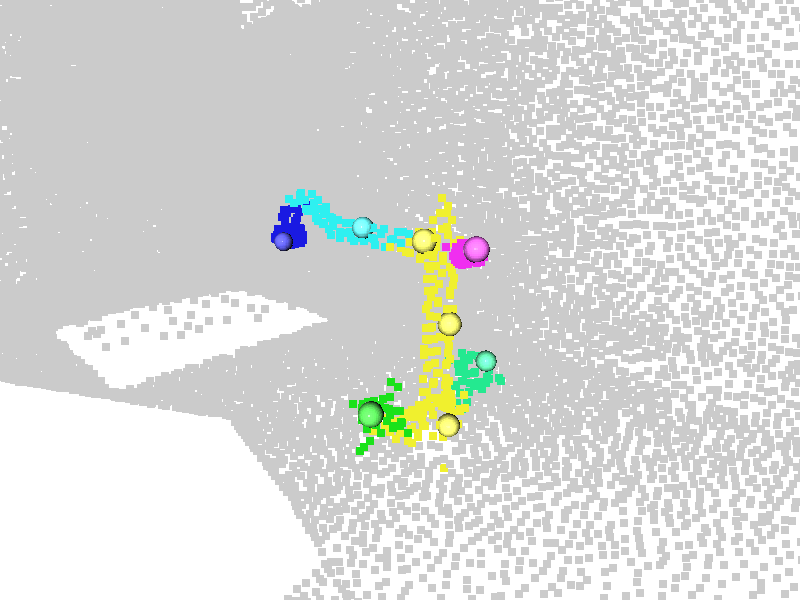} &
    \includegraphics[width=0.18\textwidth]{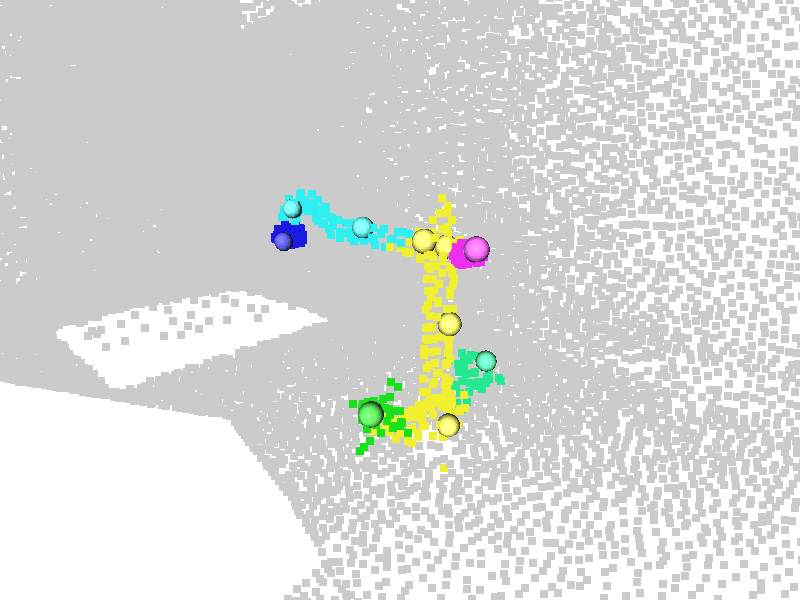} &
    \includegraphics[width=0.18\textwidth]{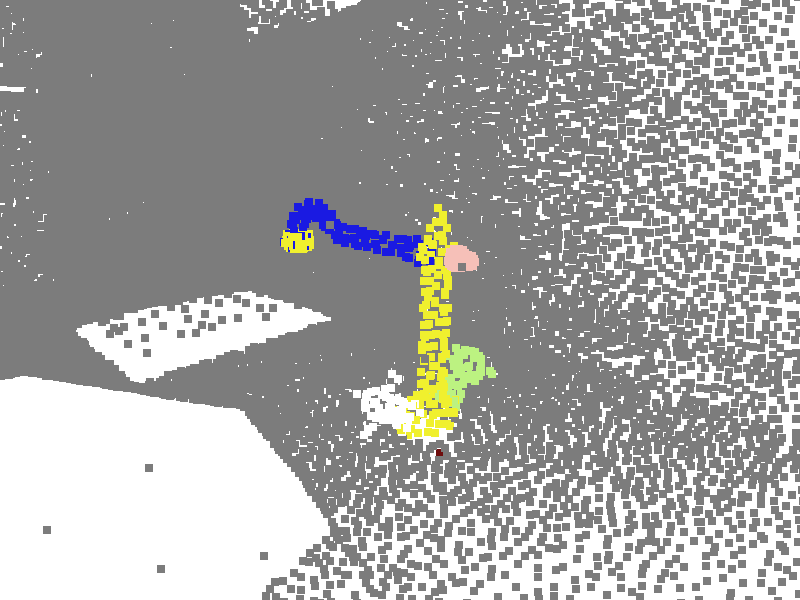} \\
    & & {\scriptsize $\overline{\text{IoU}}@6=66.0$} & {\scriptsize $\overline{\text{IoU}}@8=71.1$} & {\scriptsize $\overline{\text{IoU}}@10=75.2$} & \\
    
     &
    \includegraphics[width=0.18\textwidth]{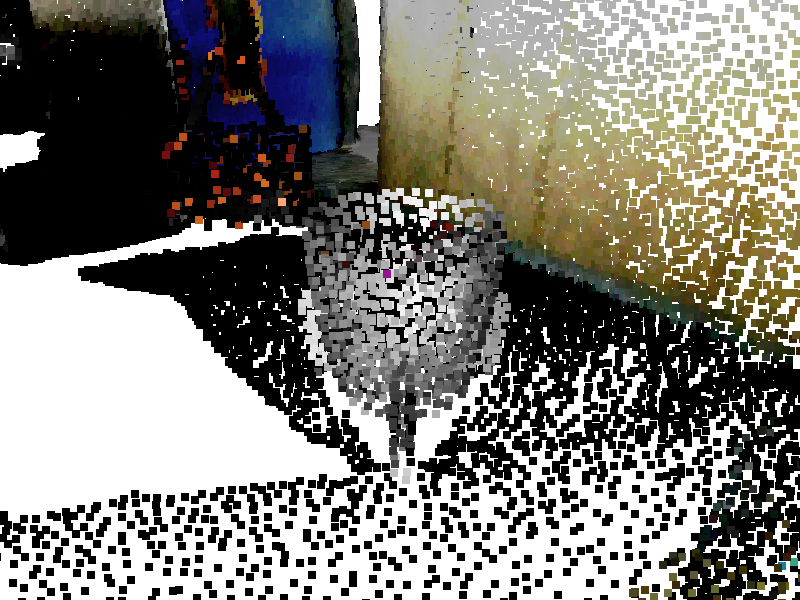} &
    \includegraphics[width=0.18\textwidth]{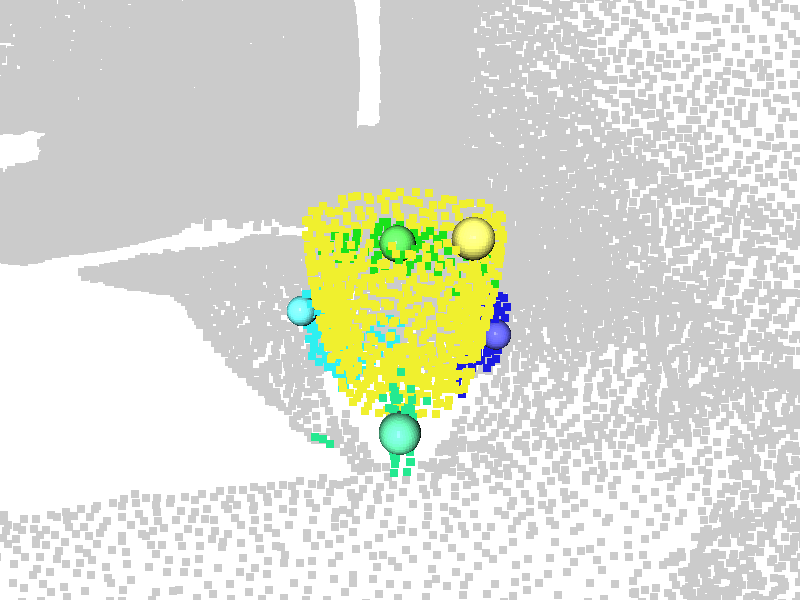} &
    \includegraphics[width=0.18\textwidth]{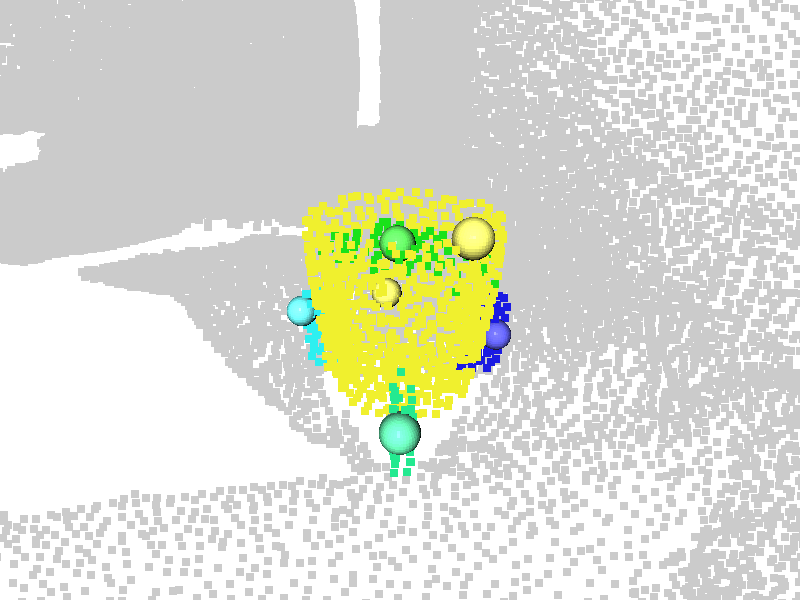} &
    \includegraphics[width=0.18\textwidth]{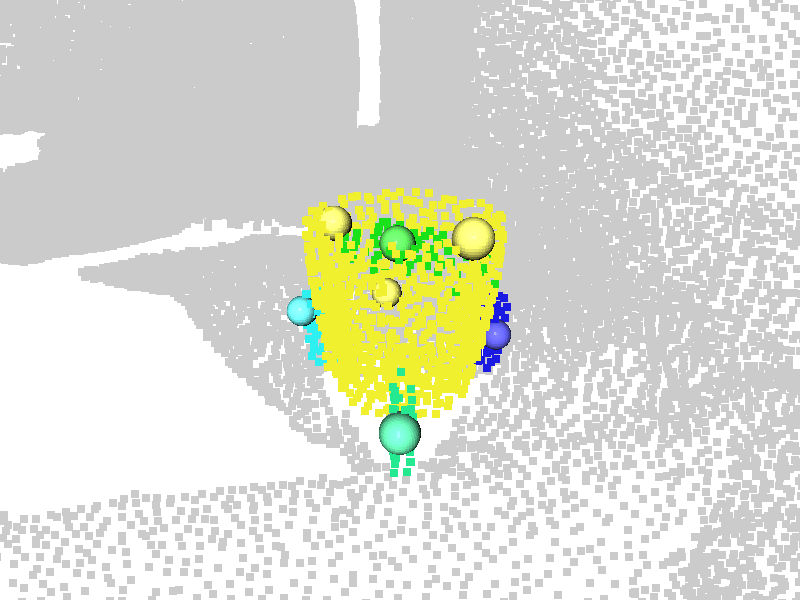} &
    \includegraphics[width=0.18\textwidth]{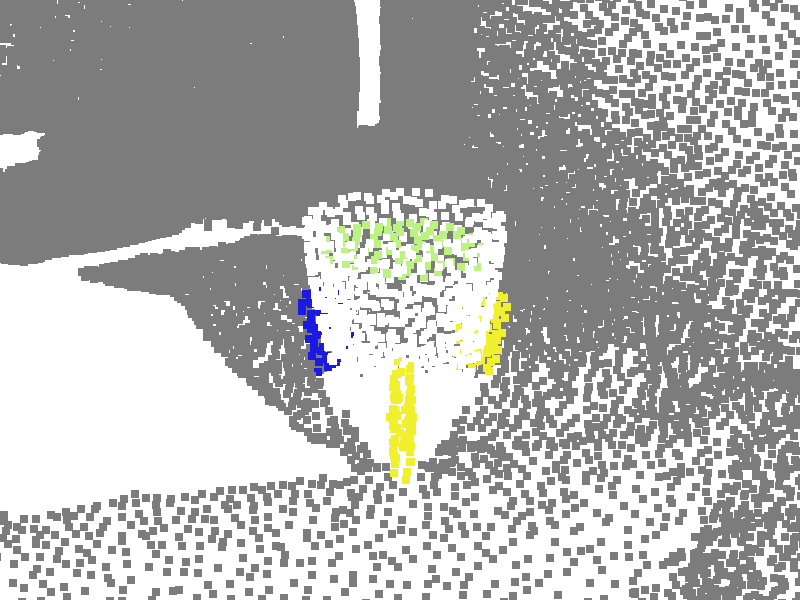} \\
    & & {\scriptsize $\overline{\text{IoU}}@5=64.3$} & {\scriptsize $\overline{\text{IoU}}@6=62.8$} & {\scriptsize $\overline{\text{IoU}}@7=64.7$} & \\

    & \multicolumn{1}{c}{3DScene} 
    & \multicolumn{3}{c}{PinPoint3D (ours)} 
    & GroundTruth \\
\end{tabular}
}

\caption{\textbf{In-domain qualitative results on PartScan.}We show interactive part segmentation on three examples from the PartScan dataset. 
Each row presents the input scene, segmentation masks produced by PinPoint3D (ours) at successive clicks, 
and the ground-truth mask.}
\label{fig:qualitative}
\end{figure*}

We extra evaluate our method on the in-domain \textbf{PartScan} benchmark, which provides complex indoor scenes with diverse object instances and fine-grained part annotations. 
As shown in Fig.~\ref{fig:qualitative}, PinPoint3D progressively refines the segmentation masks with additional user clicks, producing accurate part-level masks after only a few interactions.

In the last scene, the object contains $5$ annotated parts. 
The model reaches $\overline{\text{IoU}}=64.3$ at $5$ clicks (weighted IoU $=77.8$), $\overline{\text{IoU}}=62.8$ at $6$ clicks (weighted IoU $=78.4$), and $\overline{\text{IoU}}=64.7$ at $7$ clicks (weighted IoU $=79.5$), corresponding to about $1.4$ clicks per part. 

At click 6 the average IoU decreases slightly compared to click 5, whereas the weighted IoU continues to improve. This discrepancy arises because the average IoU assigns equal weight to every part, making it more sensitive to errors on small structures, while the weighted IoU aggregates over points and better reflects improvements on dominant regions. These results underscore the intrinsic difficulty of part-level segmentation (small, thin, or structurally complex components) and the steady gains brought by multi-click refinement.

All the cases illustrate the efficiency of our framework, requiring only a small number of interactions per part to yield satisfactory segmentation.
Overall, under in-domain conditions, our framework not only maintains accurate object-level boundaries but also delivers strong part-level segmentation efficiency, achieving high IoU scores while requiring only $1$--$2$ clicks per part on average.

\section*{MODEL DETAILS}

\subsection*{Synthetic Dataset Construction Details}
Here are the details on how to use Partfield to generate part annotations for ScanNet.

Let $F \in \mathbb{R}^{N \times C}$ denote the feature matrix of an object, 
where each row $\mathbf{f}_i \in \mathbb{R}^C$ corresponds to the feature vector 
of the $i$-th point and $N$ is the number of points. 
We perform part segmentation by clustering these point-wise features into $K$ groups 
using the K-means algorithm. Formally, the optimization objective is defined as
\begin{equation}
    \min_{\{\boldsymbol{\mu}_j\}_{j=1}^K} 
    \sum_{j=1}^K \sum_{i \in C_j} 
    \lVert \mathbf{f}_i - \boldsymbol{\mu}_j \rVert_2^2,
\end{equation}
where $\boldsymbol{\mu}_j \in \mathbb{R}^C$ is the centroid of the $j$-th cluster 
and $C_j$ is the set of points assigned to it. To automatically determine the optimal number of part clusters $k$, we compute the silhouette coefficient for each candidate $k \in \{2,\dots,10\}$. For a given point $\mathbf{p}_i$, its silhouette score $s_i$ is defined as 
\begin{equation}
s_i = \frac{b_i - a_i}{\max\{a_i,\, b_i\}}\!,
\end{equation}
where $a_i$ is the average intra-cluster distance for $\mathbf{f}_i$, and $b_i$ is the smallest average distance from $\mathbf{f}_i$ to points in any other cluster. We then select the $k$ that maximizes the average silhouette score across all points:
\begin{equation}
k^{*} = \arg\max_{k \in \{2,\dots,10\}} \frac{1}{N} \sum_{i=1}^N s_i\,. 
\end{equation}

 In practice, for each object we evaluate $k$ from 2 to 10 and choose $k^*$ as the number of parts; the points are then assigned cluster labels as their part annotations.

\subsection*{Dual-level Transformer Decoder}

\paragraph*{Click Attention}

Following AGILE3D~\cite{Yue2023}, we adopt the same four-block stack (C2S/C2C/FFN/S2C) at each pyramid level as summarized in the main text; here we only highlight implementation differences relevant to our part decoder.

In our framework, the same four-block stack is coupled with a composition-aware gating for part decoding. After object predictions are obtained, Targeted Attention Masking (TAM) constructs a memory mask over a user-specified target set $\mathcal{S}$ of instances and treats their union as a composed object. During the part stage, all cross-attention operates strictly within the union interior $\Omega(\mathcal{S})$, while background channels attend to the complement. This composition-aware restriction prevents cross-object leakage and enables part queries to specialize over fine substructures, including cases where a functional part spans multiple base instances in the scene hierarchy. Rather than functioning as a standalone module, TAM should be regarded as a strategy that integrates with both transformer stages; under joint training of the instance and part decoders, the TAM module will play a decisive role in enforcing object–part hierarchy and stabilizing refinement. Foreground queries originate from positive clicks, and background queries include a small set of learned anchors (optionally augmented by user negatives), which propagate negative context through S2C and sharpen subsequent C2S updates. Lightweight residual adapters atop the backbone projection enhance part-level sensitivity without discarding object-level geometry, and temporal encodings make the refinement order-aware while preserving permutation invariance via late max fusion of per-point logits within the same semantic group.


\paragraph*{Implementation-level Aspects and Additional Observations on Targeted Attention Masking}

In practice, the mask is recomputed at every refinement step using the current per-point predictions of object versus background, ensuring that query-point interactions always align with the latest predictions. Background queries are explicitly constrained to background-only points, which suppresses leakage into neighboring objects and improves separation at instance boundaries. Although the theoretical complexity of attention remains $O(QN)$, restricting keys to object interiors reduces the effective attention domain in practice, leading to lower runtime and memory consumption in multi-object scenes.

Beyond these mechanics, multi-object training yields characteristic behaviors that complement the high-level factors summarized in the main paper: dispersed positive clicks act as an implicit regularizer by encouraging the decoder to explain each click in place rather than assuming a single connected region; the need to correct over-merged object masks transfers into more conservative growth and cleaner boundaries at the part level; and dispersed supervision reduces inter-query coupling, so queries compete softly and specialize as local experts over compact regions.

\subsection*{Implementation Details}

\paragraph*{Backbone and adapters}
A sparse backbone produces per-point features that are projected to the decoder width. A lightweight residual adapter specializes features for part-level decoding while preserving object-level semantics; residual scaling (e.g., $\alpha\!\in\!(0,1]$) and careful initialization improve stability. By default, the adapter is the only trainable component in the backbone; other weights remain frozen unless stated.

\paragraph*{Queries and fusion}
\textbf{Foreground \& background queries.} Foreground queries are derived from user clicks; background queries comprise user-provided negatives (when available) plus a small set of learned background anchors that improve coverage in sparsely clicked scenes.
\textbf{Late fusion rationale.} Within each semantic group (foreground objects, background, or part channels), logits from multiple queries are merged by a per-point max. This yields permutation-invariant label assignment with respect to the click order, avoids averaging-induced shrinkage, and encourages complementary specialization among queries.

\paragraph*{Instance-to-part handoff}

\textbf{Invocation.} The part stage is invoked for the user-selected instance (or composed object); its queries attend only to the instance interior defined by TAM, while background queries remain confined to background regions.
\textbf{Iterative refinement.} When new clicks arrive, temporal indices are appended and the part stage is rerun with a refreshed TAM gate derived from the latest object logits; both query states and masks are refined without re-encoding the whole scene.

\paragraph*{Numerical stability and precision}
\textbf{Mixed precision.} Positional encodings are computed in full precision to avoid overflow in high-frequency bands; attention and feed-forward blocks use mixed precision for efficiency.  
\textbf{Mask smoothing (optional).} While we use hard labels for TAM, temperature-scaled soft gates are compatible and can be used to trade boundary crispness for smoother gradients when desired.

\paragraph*{Hierarchy and pyramids}
\textbf{Pyramid scheduling.} Both instance and part stages operate over a feature pyramid; each layer's S2C write-back makes the next resolution level aware of the latest query context. The TAM gate itself is \emph{resolution-agnostic} and is applied consistently across levels.

\paragraph*{Typical configuration (non-exhaustive)}
Hidden size, head count, feed-forward expansion, and the number of decoder layers follow standard transformer practice; weights can be shared across layers or left independent depending on memory constraints. Background anchors are few in number; their presence is consistent across scenes and does not depend on user input.

\paragraph*{Engineering notes for reproducibility}
\textbf{Click batching.} Clicks are grouped per interaction step; the temporal index resets between sessions but remains strictly increasing within a session.  
\textbf{Empty/intersecting masks.} If an object mask becomes empty after a refinement step, TAM falls back to the previous non-empty interior to avoid stalling. Overlaps between object interiors are resolved by the instance argmax before TAM is formed.  
\textbf{Determinism.} With a fixed random seed, normalization range and nearest-neighbor fallbacks make click-to-query initialization deterministic across runs.

\subsection*{Training Details}

We adopt a progressive multi-step interactive training paradigm, simulating 0--19 rounds of user interactions per episode. Only the final refinement step contributes gradients, while intermediate steps are used to mimic the iterative annotation process. Both positive and negative clicks are supported: positives are sampled on mis-segmented regions of the target part, whereas negatives are drawn from background areas. To better reflect realistic scene-level behavior, the number of involved parts is adaptively selected 3--10 rather than fixed.

During validation, we evaluate under three modes: (i) \emph{single-part}, where only one target part is segmented; (ii) \emph{multi-part}, where segmentation is conditioned on a designated object; and (iii) \emph{full-part}, where all annotated parts are considered.

Loss supervision combines binary cross-entropy and Dice objectives applied to both object- and part-level predictions, with default coefficients $\lambda_{\text{CE}}{=}1.0$ and $\lambda_{\text{Dice}}{=}2.0$. Auxiliary supervision from intermediate decoder layers is enabled by default. Additionally, click-adaptive weighting is used so that errors around clicked regions are penalized more heavily.  

Optimization uses AdamW with an initial learning rate of $1\times10^{-4}$, weight decay of $1\times10^{-4}$, and gradient clipping at $0.1$. A multi-step schedule decays the learning rate after 1000 epochs. Training runs for 1100 epochs in total, with validation performed every 50 epochs. Batch sizes are 5 for training and 1 for validation.  

To preserve stable object-level semantics, the backbone and object-level decoder remain frozen. A lightweight adapter (bottleneck ratio 1.0) is trained jointly with the part-level decoder, while cross-attention modules (part$\to$object, part self-attention, etc.) are updated to capture fine-grained semantics.

\section*{USER STUDY}

\begin{figure*}[t!]
\centering

\begin{minipage}{0.8\textwidth}
    \centering
    \includegraphics[width=\linewidth]{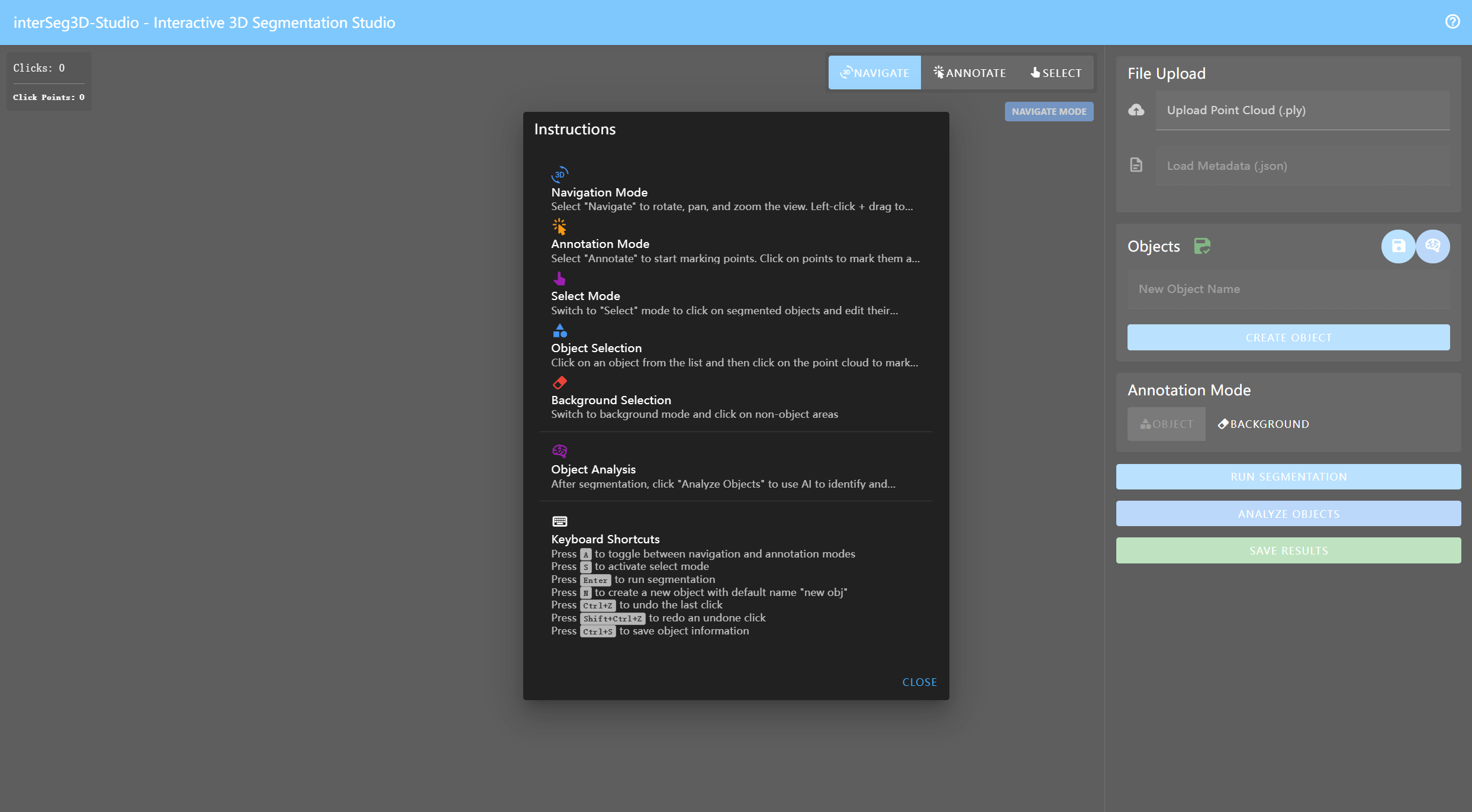}
    \\\small Frontend tutorial page
\end{minipage}

\vspace{6pt} %

\begin{minipage}{0.8\textwidth}
    \centering
    \includegraphics[width=\linewidth]{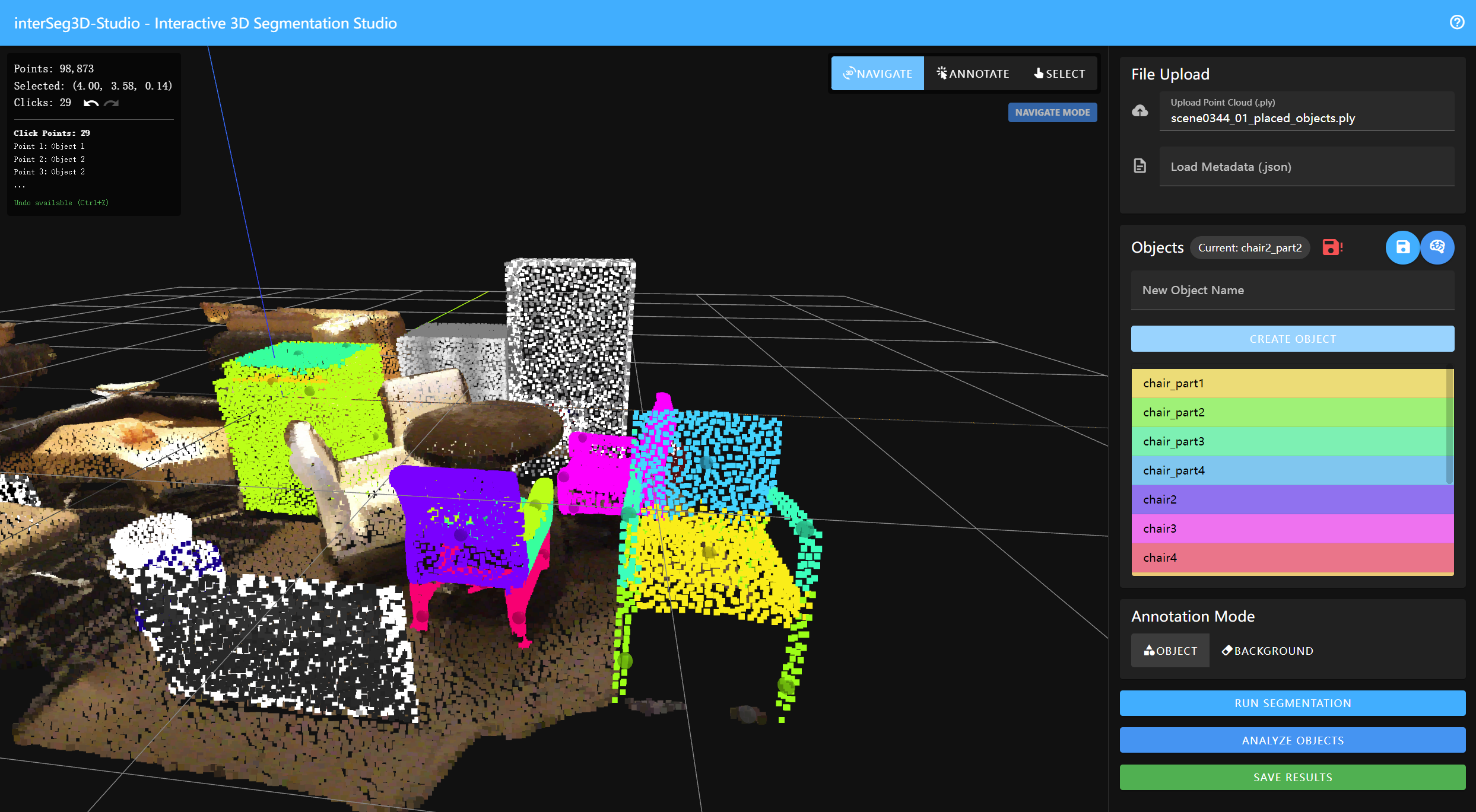}
    \\\small Actual user interface of our system
\end{minipage}

\caption{\textbf{Annotation interface for the user study.}}

\label{fig:ui}
\end{figure*}

\subsection*{Setting}

\begin{table}[t!]
\centering
\small
\caption{Comparison between Human and Simulation Users}
\label{tab:user-study}
\renewcommand{\arraystretch}{1.2}
\setlength{\tabcolsep}{10pt}
\begin{tabular}{llcccccc}
\toprule
User & Dataset & $\overline{\text{IoU}@3}\uparrow$  & $\overline{\text{IoU}@5}\uparrow$  \\
\midrule
Human & \multirow{2}{*}{PartScan} & \textbf{61.9} & \textbf{74.7}  \\
Sim.  &                            & 54.7 & 73.7     \\
\midrule
Human & \multirow{2}{*}{MultiScan} & \textbf{64.5}& 73.7 \\
Sim.  &                         & 55.1 & \textbf{74.3 }   \\
\bottomrule
\end{tabular}
\end{table}

To evaluate the effectiveness of our interactive segmentation framework under real human interactions, we conducted a user study with six participants. All participants were non-experts and had not used similar tools before. Each user was first provided with written instructions and a verbal demonstration of the annotation interface, followed by a 15 minute practice session on example scenes. After familiarization, users were asked to complete fine-grained part segmentation on 20 scene point clouds, where each scene contained 5--10 parts to be segmented. 
Table~\ref{tab:user-study} shows that human annotators achieve comparable results to the simulated users. Furthermore, user studies on Multiscan illustrate that our method generalizes well on unseen data, reaching similar results as on our training sets. Details can be found in the Appendix.

\subsection*{User Interface}
Our annotation interface (Fig.~\ref{fig:ui}) is implemented with Vue.js and Three.js, providing a lightweight web-based platform that runs across operating systems and browsers. It supports both single- and multi-object segmentation through intuitive click-based interactions. Users can switch between navigation, annotation, and selection modes, with keyboard shortcuts (e.g., \emph{A} to toggle navigation/annotation, \emph{S} for selection, Ctrl+Z for undo) to streamline the workflow. The interface offers visual feedback via color-coded highlighting and click statistics, and automatically preserves annotation states to prevent data loss. We will release the interface together with the source code to facilitate future research in interactive 3D segmentation.

\begin{figure}[t]
    \centering
    \includegraphics[width=0.9\linewidth]{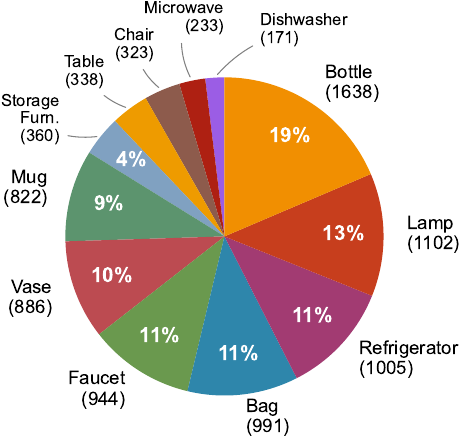}
    \caption{\textbf{Categories of objects in the training datasets.} }
    \label{fig:part-category}
\end{figure}

\begin{figure}[t]
    \centering
    \includegraphics[width=0.9\linewidth]{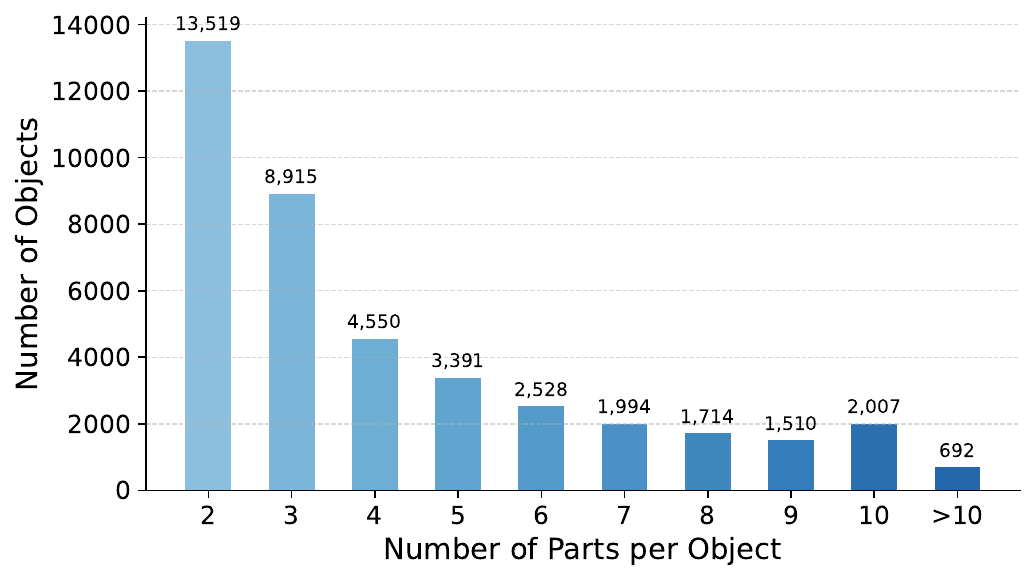}
    \caption{\textbf{Part distribution in PartScan.}}
    \label{fig:part-distribution}
\end{figure}

\section*{BENCHMARK DETAILS}
\subsection*{Dataset}

\textbf{ScanNet}~\cite{dai2017ScanNet} is a richly-annotated dataset of 3D indoor scenes, covering diverse room types such as offices, hotels, and living rooms. The dataset provides dense semantic segmentation masks for 40 object classes. To test model generalization, we train on the ScanNet20 split and evaluate on both the 20 seen classes and the 20 unseen classes. In our interactive segmentation setup, we simulate user-provided clicks to iteratively segment target objects within the ScanNet scenes.

\textbf{MultiScan}~\cite{mao2022multiscan} is a scalable RGB-D dataset of indoor environments with articulated objects. The dataset comprises 273 scans of 117 scenes containing 10,957 object instances and 5,129 part instances, all captured with a commodity RGB-D sensor and reconstructed into textured 3D meshes. Each scan provides rich semantic annotations at both the object and part levels, as well as annotated part mobility parameters. We use MultiScan to benchmark interactive segmentation of both object instances and their constituent parts.

\textbf{PartNet}~\cite{mo2019partnet} is a large-scale dataset of 3D objects with fine-grained, hierarchical part annotations. It contains 26,671 models across 24 object categories, with a total of 573,585 annotated part instances. The dataset supports multiple levels of part hierarchy, enabling evaluation of tasks such as fine-grained semantic part segmentation and hierarchical part segmentation. We integrate 3D models from PartNet into scene scans to generate datasets for training and evaluations.

\textbf{KITTI-360}~\cite{kitti-360} is a large-scale outdoor driving dataset with synchronized LiDAR and camera recordings. It contains 100k LiDAR scans collected over long driving trajectories. For our experiments, we focus on interactive 3D segmentation in one representative sequence (2013\_05\_28\_drive\_0000\_sync). 

\subsection*{Our dataset}
We constructed a synthesized dataset by integrating PartNet assets into scenes. 12 object categories from PartNet, including \textit{Table, Refrigerator, StorageFurniture, Chair, Dishwasher, Microwave, Bag, Mug, Bottle, Lamp, Vase, Faucet}, are selected based on three primary criteria: (1) high prevalence in indoor scenes, (2) scale compatibility with ScanNet environments, and (3) possession of distinct, semantically meaningful part structures. Tab.~\ref{tab:per-category-iou} reports IoU achieved by PinPoint3D across the selected categories. These results highlight both the diversity of categories and the variation in segmentation difficulty across object types. Fig.~\ref{fig:part-category} and Fig.~\ref{fig:part-distribution} summarize the number of instances in each category and the part count distribution of our constructed dataset, highlighting the diversity of structural complexity across objects.

\end{document}